\pdfoutput=1

\documentclass[11pt]{article}

\usepackage[final]{acl}

\usepackage{times}
\usepackage{latexsym}

\usepackage[T1]{fontenc}

\usepackage[utf8]{inputenc}

\usepackage{microtype}

\usepackage{inconsolata}

\usepackage{graphicx}

\usepackage{amsmath}
\usepackage{amsfonts}
\usepackage{graphicx}
\usepackage{float}
\usepackage{enumitem}
\usepackage{mathtools}
\usepackage{multirow}
\usepackage{tabularx}
\usepackage{booktabs}

\usepackage[toc,page,header]{appendix}
\usepackage{minitoc}

\title{ThinkEdit: Interpretable Weight Editing to \\ Mitigate Overly Short Thinking in Reasoning Models}

\author{Chung-En Sun \\
  UCSD CSE \\
  \texttt{cesun@ucsd.edu} \\
  \And
  Ge Yan \\
  UCSD CSE \\
  \texttt{geyan@ucsd.edu} \\
  \And
  Tsui-Wei Weng \\
  UCSD HDSI \\
  \texttt{lweng@ucsd.edu} \\
  }

\begin{document}
\maketitle
\begin{abstract}
Recent studies have shown that Large Language Models (LLMs) augmented with chain-of-thought (CoT) reasoning demonstrate impressive problem-solving abilities. However, in this work, we identify a recurring issue where these models occasionally generate overly short reasoning, leading to degraded performance on even simple mathematical problems. Specifically, we investigate how reasoning length is embedded in the hidden representations of reasoning models and its impact on accuracy. Our analysis reveals that reasoning length is governed by a linear direction in the representation space, allowing us to induce overly short reasoning by steering the model along this direction. Building on this insight, we introduce \textbf{\textit{ThinkEdit}}, a simple yet effective weight-editing approach to mitigate the issue of overly short reasoning. We first identify a small subset of attention heads (approximately 4\%) that predominantly drive short reasoning behavior. We then edit the output projection weights of these heads to remove the short reasoning direction. With changes to only 0.2\% of the model's parameters, \textbf{\textit{ThinkEdit}} effectively reduces overly short reasoning and yields notable accuracy gains for short reasoning outputs (+6.39\%), along with an overall improvement across multiple math benchmarks (+3.34\%). Our findings provide new mechanistic insights into how reasoning length is controlled within LLMs and highlight the potential of fine-grained model interventions to improve reasoning quality. Our code is available at: \textsf{{\small \href{https://github.com/Trustworthy-ML-Lab/ThinkEdit}{https://github.com/Trustworthy-ML-Lab/ThinkEdit}}}
\end{abstract}

\doparttoc 
\faketableofcontents 

\section{Introduction}
\label{sec:introduction}
Recently, Reinforcement Learning (RL) has been applied to enhance Large Language Models (LLMs), equipping them with strong chain-of-thought (CoT) reasoning abilities \cite{r1}. These models, often referred to as reasoning models, first generate an intermediate reasoning process -- a "thinking step" -- where they reason step-by-step and then self-correct before producing a final response. As a result, they achieve remarkable improvement on mathematical reasoning tasks and demonstrate a strong ability to generate detailed CoT reasoning \cite{o1, r1, s1}.

However, despite these improvements, reasoning models still exhibit a non-negligible gap from perfect accuracy on relatively simple benchmarks such as GSM8K \cite{gsm8k}. As shown in Section~\ref{sec:short_reasoning_prevalence}, we found that Deepseek-distilled reasoning models occasionally generate overly short reasoning chains, which correlate with lower accuracy (about 20\% drop on MATH-level5 benchmark \cite{math}). This issue appears consistently across models of different sizes, suggesting that reasoning length plays a crucial role in problem-solving effectiveness. Yet, the mechanisms governing reasoning length within the model’s internal representation remain underexplored, despite being crucial for understanding reasoning models.

\begin{figure*}[t]
    \centering
    \includegraphics[width=1.0\textwidth]{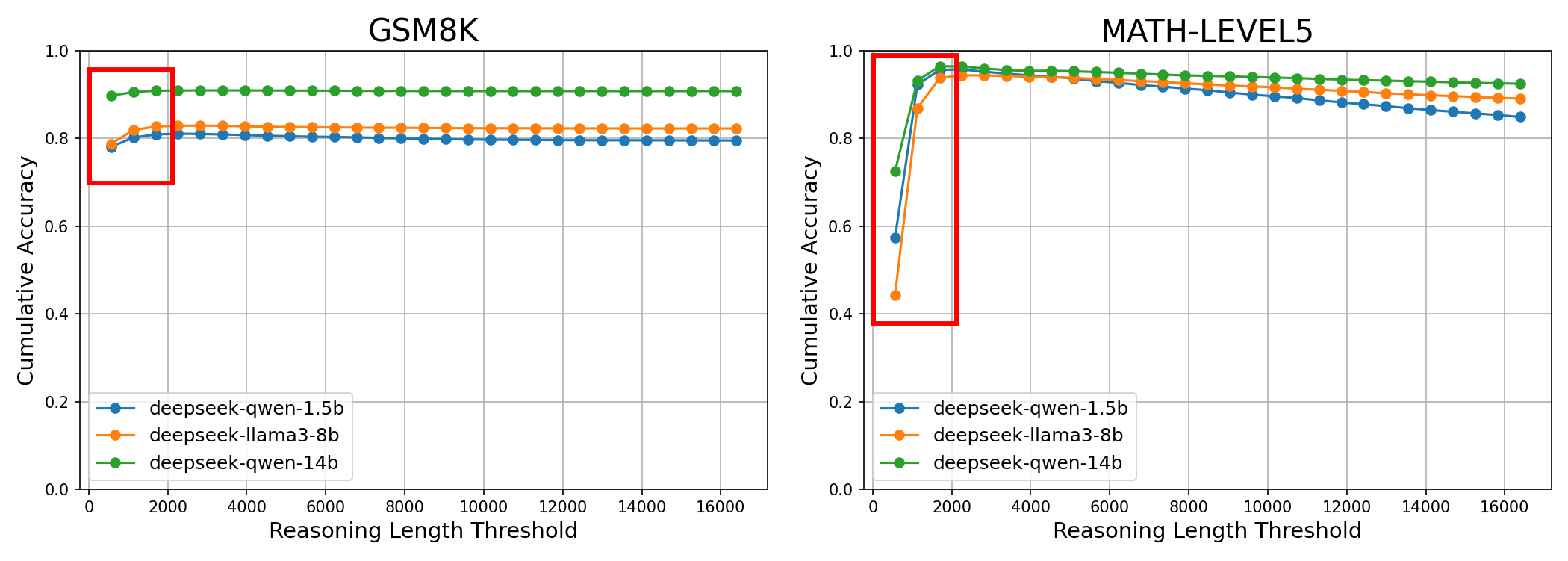}
    \vspace{-20pt}
    \caption{Cumulative accuracy as a function of the reasoning length threshold. The x-axis represents the cutoff threshold on reasoning length, and the y-axis shows the average accuracy of all responses with reasoning length below that threshold. Models consistently exhibit lower accuracy for overly short reasoning (e.g. length <1000).}
    \label{fig:acc_vs_threshold}
\end{figure*}

\begin{figure*}[t]
    \centering
    \includegraphics[width=1.0\textwidth]{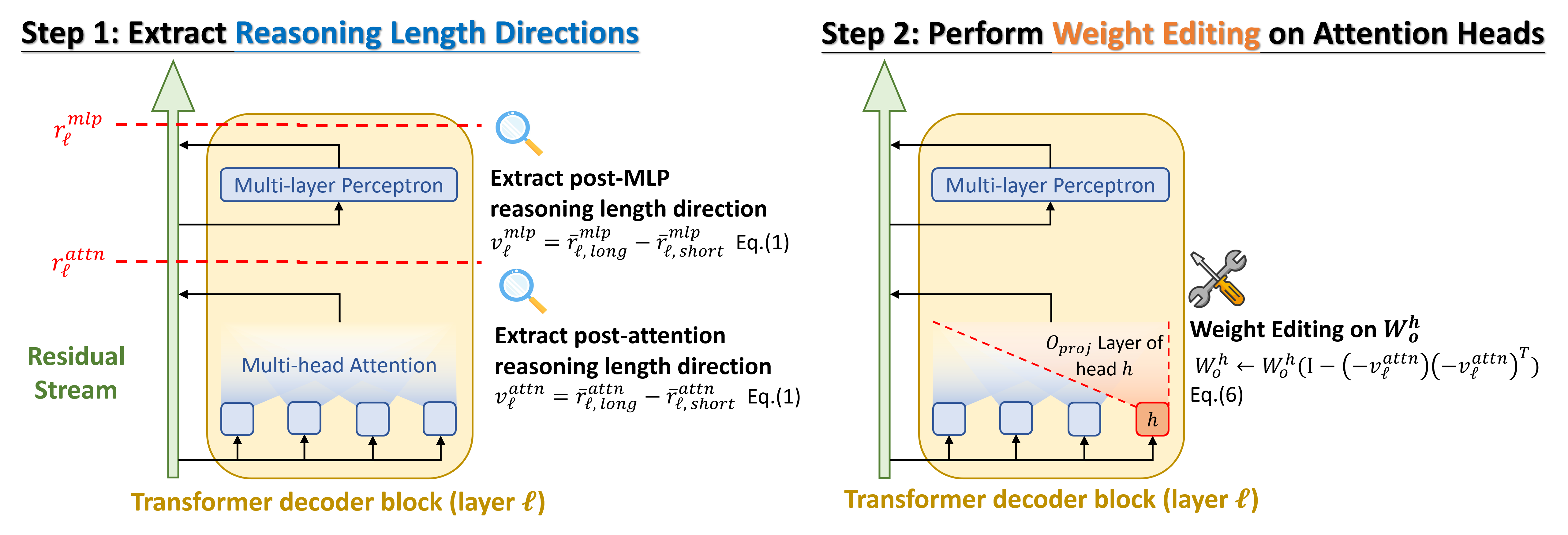}
    \vspace{-20pt}
    \caption{The overview of \textbf{\textit{ThinkEdit}} framework. We first identify that there exist linear directions for controlling reasoning length in the hidden space, and then perform weight editing on the key attention heads.}
    \label{fig:overview}
\end{figure*}

To bridge this gap, in this work, we first investigate how reasoning length is encoded within the hidden representations of reasoning models. By performing a novel analysis of the residual stream, we extract a \textit{reasoning length direction}—a latent linear representation in the residual stream that enables direct control over reasoning length as shown in Figure~\ref{fig:overview} (left). Our analysis reveals that overly short, abstract, or high-level reasoning significantly degrades model performance, and this characteristic is primarily embedded in the middle layers of the model. Furthermore, we identify a small subset (approximately 4\%) of attention heads in the middle layers that disproportionately contribute to short reasoning. Building on this insight, we propose \textbf{\textit{ThinkEdit}}, a simple and effective weight-editing technique to remove the short-reasoning component from these attention heads' output projection layers, as shown in Figure ~\ref{fig:overview} (right). Our findings demonstrate that disabling these components leads to a non-trivial improvement in accuracy when the model generates short reasoning while also enhancing overall performance. Our contributions are summarized as follows:
\vspace{10pt}
\begin{itemize}[leftmargin=0.5cm]
    \item We identify the prevalence of overly short reasoning across Deepseek-distilled reasoning models of different scales and highlight its impact on the performance of math benchmarks.
    \item We extract a \textit{reasoning length direction} in the model’s hidden representations, revealing that \textbf{middle layers} play a crucial role in controlling reasoning length. To the best of our knowledge, this is the first work to systematically study the internal representations of reasoning models.
    \item We discover a small set of \textbf{"short reasoning" heads} that strongly contribute to the generation of brief reasoning chains and propose \textbf{\textit{ThinkEdit}}. By editing the output projection weights of just 4\% heads (0.2\% of the model’s total parameters), \textbf{\textit{ThinkEdit}} effectively mitigates short reasoning, leading to improved performance both when short reasoning occurs (+6.39\%) and in overall accuracy (+3.34\%).
\end{itemize}
\vspace{15pt}
\section{Unexpectedly Low Accuracy in Short Reasoning Cases}
\label{sec:short_reasoning_prevalence}

We begin our study by highlighting a consistent issue observed in Deepseek-distilled reasoning models across a variety of sizes: significantly lower accuracy when the reasoning length is short. This pattern holds across datasets such as GSM8K \cite{gsm8k} and MATH-Level5 \cite{math}. Figure~\ref{fig:acc_vs_threshold} illustrates this trend, with the x-axis indicating a cutoff threshold on reasoning length. For example, a threshold of 2000 denotes that we calculate the average accuracy over all responses whose reasoning length is at most 2000 tokens. The y-axis shows the corresponding cumulative accuracy. The details of the experimental setup are provided in Section~\ref{sec:perf_edited_models}.

Contrary to intuition, one might expect shorter reasoning to correspond to easier questions, as such problems should require fewer steps to solve. This expectation is partially supported by the trend in Figure~\ref{fig:acc_vs_threshold} (right), where accuracy tends to decrease as reasoning length exceeds 2000. However, the region with reasoning length below 2000 (highlighted in red boxes) exhibits a different pattern: models consistently underperform on these short-reasoning cases, with accuracy dropping significantly below the overall average. This suggests that, rather than efficiently solving simple problems with brief reasoning, models often fail when producing overly short chains of thought.

Motivated by this observation, we focus on investigating how a model’s internal representations govern reasoning length and influence accuracy. In Section~\ref{sec:understanding_reasoning_length}, we analyze the relationship between hidden representations, reasoning length, and model performance. Building on these insights, we propose \textbf{\textit{ThinkEdit}}, a simple yet effective weight-editing method, in Section~\ref{sec:mitigate_short_reasoning}, which modifies the output layer of a few key attention heads to mitigate the problem of overly short reasoning.

\section{Understanding How Representations Affect Reasoning Length}
\label{sec:understanding_reasoning_length}
In this section, we explore how reasoning length is encoded in the hidden representation of a reasoning model. In Section \ref{sec:transformer_preliminaries}, we provide an overview of the transformer structure, highlighting the specific points in the residual stream where the representation of interest resides. Then, in Section \ref{sec:extract_reasoning_length_directions}, we present our method for extracting linear directions that allow control over reasoning length. Finally, in Section \ref{sec:effects_of_reasoning_length_direction}, we analyze the performance of reasoning models when guided by these extracted reasoning length directions.

\subsection{Background of Transformer Structure and Notations}
\label{sec:transformer_preliminaries}

A transformer model consists of multiple stacked layers, each containing a \textit{multi-headed self-attention} (Attn) module followed by a \textit{feed-forward Multi-Layer Perceptron} (MLP). The model maintains an evolving \textit{Residual Stream}, where representations are progressively refined as they pass through layers. The update at each layer $\ell$ can be expressed as:
\vspace{-5pt}
\[
\begin{aligned}
r^{\text{attn}}_\ell &= r^{\text{mlp}}_{\ell-1} + \textrm{Attn}(\textrm{LayerNorm}(r^{\text{mlp}}_{\ell-1})) \\
r^{\text{mlp}}_{\ell} &= r^{\text{attn}}_\ell + \textrm{MLP}(\textrm{LayerNorm}(r^{\text{attn}}_\ell))
\end{aligned}
\vspace{-5pt}
\]
where $r^{\text{mlp}}_{\ell-1}$ is the hidden state entering layer $\ell$, which is also the output of the MLP from the previous layer $\ell-1$, $r^{\text{attn}}_\ell$ represents the intermediate state of the residual stream after the self-attention module, and $r^{\text{mlp}}_{\ell}$ denotes the final output after the MLP transformation.  

Our focus is on the hidden representations $r^{\text{attn}}_\ell$ and $r^{\text{mlp}}_{\ell}$ as illustrated in Figure~\ref{fig:overview} (left), which capture the model's state after the self-attention and MLP transformations, respectively.

\subsection{Extracting Reasoning Length Directions}
\label{sec:extract_reasoning_length_directions}

To investigate how \emph{reasoning length} is encoded in a model's hidden representation, we begin by collecting the model’s responses to 2{,}000 problems from the GSM8K \cite{gsm8k} training set. In each response, the chain-of-thought (CoT) is enclosed between special tags \verb|<think>| and \verb|</think>|. We measure the length of each CoT by counting only the tokens within these tags. We then construct two datasets $\mathcal{D}_{\text{long}}$ and $\mathcal{D}_{\text{short}}$, where $\mathcal{D}_{\text{long}}$ consists of responses whose CoT exceeds 1000 tokens and $\mathcal{D}_{\text{short}}$ includes those under 100 tokens. Each entry in these datasets contains: (1) the problem statement, (2) the extracted CoT, enclosed by \verb|<think>| and \verb|</think>| tags, and (3) the step-by-step calculation process leading to the final answer.

Next, we input the problem statement along with its CoT into the model and extract hidden representations at each layer \(\ell\) for both the \emph{post-attention} and \emph{post-MLP} residual streams, denoted as \( r^{\text{attn}}_{\ell} \) and \( r^{\text{mlp}}_{\ell} \), respectively. Specifically, let \( r^{\text{attn}}_{\ell}(i, t) \) and \( r^{\text{mlp}}_{\ell}(i, t) \) represent the hidden representations at layer \(\ell\) for token position \(t\) in the response to problem \(i\). We first compute the mean hidden representation over the chain-of-thought (CoT) tokens, where \(\mathcal{T}_i\) denotes the set of token positions enclosed within the \verb|<think>| and \verb|</think>| tags, and then compute the mean across all problems in the datasets \(\mathcal{D}_{\text{long}}\) and \(\mathcal{D}_{\text{short}}\), yielding layerwise embeddings:
\vspace{-5pt}
\[
\overline{r}^{x}_{\ell, y} = \frac{1}{|\mathcal{D}_y|} \sum_{i \in \mathcal{D}_y} \frac{1}{|\mathcal{T}_i|} \sum_{t \in \mathcal{T}_i} r^{x}_{\ell}(i, t),
\vspace{-5pt}
\]
where \(x \in \{\text{attn}, \text{mlp}\}\) denotes the representation type and \(y \in \{\text{long}, \text{short}\}\) indicates the reasoning-length group.
Finally, we define the \emph{reasoning-length direction} at layer~$\ell$ as the vector difference between the “long” and “short” embeddings:
\begin{equation}
\label{e:reasoning_length_direction}
v^{\text{attn}}_{\ell} = \overline{r}^{\text{attn}}_{\ell,\text{long}} - \overline{r}^{\text{attn}}_{\ell,\text{short}}, \;\;
v^{\text{mlp}}_{\ell} = \overline{r}^{\text{mlp}}_{\ell,\text{long}} - \overline{r}^{\text{mlp}}_{\ell,\text{short}}.
\end{equation}
These two vectors, $v^{\text{attn}}_{\ell}, v^{\text{mlp}}_{\ell} \in \mathbb{R}^d$ (with $d$ denoting the hidden dimension), capture how the model’s representation differs when reasoning chains are notably longer or shorter. In the next section, we analyze how modifying these directions in the residual stream influences both reasoning length and overall model performance.

\begin{figure*}[!t]
    \centering
    \includegraphics[width=1.0\textwidth]{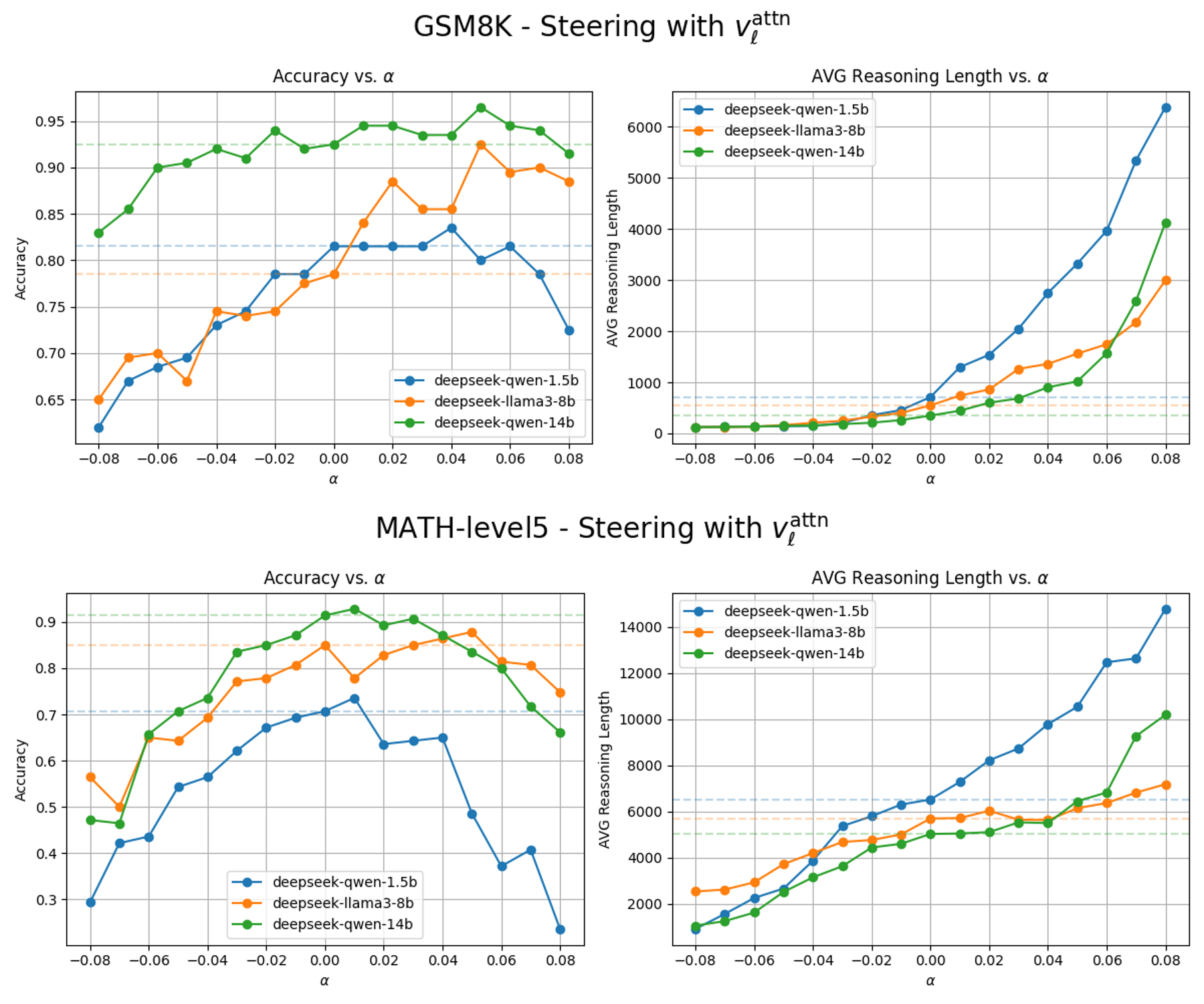}
    \vspace{-20pt}
    \caption{
        Global steering results. \textbf{Top:} On GSM8K, positive \(\alpha\) extends reasoning length and improves accuracy in the 8B and 14B models, while negative \(\alpha\) shortens reasoning and lowers accuracy. \textbf{Bottom:} On MATH-Level5, negative \(\alpha\) similarly shortens reasoning and reduces accuracy.
    }
    \label{fig:steering}
\end{figure*}

\subsection{Effects of Reasoning-Length Direction}
\label{sec:effects_of_reasoning_length_direction}

In Section~\ref{sec:extract_reasoning_length_directions}, we have obtained the steering vectors \(v^{\text{attn}}_{\ell}\) and \(v^{\text{mlp}}_{\ell}\) for reasoning length. We now investigate how modifying the residual stream along these directions affects both reasoning length and model accuracy. We begin with \emph{global steering}, where we apply a uniform shift \(\alpha\) across all layers, and then delve into \emph{layerwise} steering experiments to locate the portions of the network most responsible for reasoning length.

\paragraph{Steering Reasoning Models with \(v^{\text{attn}}_{\ell}\) and \(v^{\text{mlp}}_{\ell}\).}
Let \(\alpha\) be a scalar weight in the range \([-0.08,\, 0.08]\). For each layer \(\ell\), we apply the following transformations:
\vspace{-5pt}
\begin{equation}
\label{eq:steering_formula}
r^{\text{attn}}_{\ell}\leftarrow r^{\text{attn}}_{\ell} +  \alpha\,v^{\text{attn}}_{\ell},
\quad
r^{\text{mlp}}_{\ell}\leftarrow r^{\text{mlp}}_{\ell} + \alpha\,v^{\text{mlp}}_{\ell}.
\vspace{-5pt}
\end{equation}
This operation \emph{steers} the model’s internal states either \emph{toward longer reasoning} (if \(\alpha>0\)) or \emph{toward shorter reasoning} (if \(\alpha<0\)).

\paragraph{Experimental Setup.}
We evaluate the effect of reasoning-length directions using two test sets:
\begin{itemize}[leftmargin=0.5cm]
    \item \textbf{GSM8K (200 problems) \cite{gsm8k}:} A simpler benchmark, consisting of the first 200 problems from the GSM8K test set.
    \item \textbf{MATH-Level5 (140 problems) \cite{math}:} A more challenging benchmark, comprising 140 problems selected from the MATH test set. Specifically, we extract 20 level-5 examples from each of 7 categories.
\end{itemize}
We set a maximum reasoning length of 8,192 tokens for GSM8K and 16,384 tokens for MATH-Level5. Upon reaching this limit, the model is prompted to finalize its answer immediately. We experiment on three reasoning models of varying sizes: \texttt{deepseek-distill-qwen-1.5B}, \texttt{deepseek-distill-llama3-8B}, and \texttt{deepseek-\\distill-qwen-14B}.

\paragraph{Global Steering on GSM8K and MATH-Level5.}
Figure~\ref{fig:steering} (Top) shows the effect of applying the attention-based direction \(v_{\ell}^{\text{attn}}\) on GSM8K. We vary \(\alpha\) from \(-0.08\) (shorter reasoning) to \(+0.08\) (longer reasoning). Across all models, \emph{increasing} \(\alpha\) extends the length of CoT (Figure~\ref{fig:steering}, top right), indicating that \(v_{\ell}^{\text{attn}}\) indeed encode \textit{reasoning-length} attributes. In terms of accuracy, the larger 8B and 14B models improve when steered toward longer reasoning—particularly \texttt{deepseek-distill-llama3-8B} \textcolor{orange}{(orange line)}, which benefits most from positive steering with 10\% accuracy improvement. In contrast, the smaller \texttt{deepseek-distill-qwen-1.5B} \textcolor{blue}{(blue line)} model experiences a 10\% drop in accuracy. Figure~\ref{fig:steering} (Bottom) presents the results for the more challenging MATH-Level5 dataset. Similar to GSM8K, our extracted directions effectively control reasoning length as expected, with negative \(\alpha\) consistently leading to shorter CoT and positive \(\alpha\) extending them. In terms of accuracy, shorter reasoning also consistently degrades performance. However, unlike GSM8K, there is no clear trend indicating that longer reasoning reliably enhances accuracy; while moderate positive \(\alpha\) might provide some benefits, excessively long reasoning often negatively impacts performance. We also present results using the MLP-based direction \(v_{\ell}^{\text{mlp}}\) in Appendix~\ref{sec:mlp_steering}, which exhibit similar trends.

\paragraph{Layerwise Steering Analysis.} We perform a \emph{layerwise} experiment to identify which layers produce reasoning-length directions with the strongest impact. As shown in Appendix~\ref{sec:layerwise_steering}, \emph{middle layers} are most effective, suggesting their key role in encoding reasoning-length representations.

\paragraph{Budget Control with Steering Representations.}  
Recent work \cite{s1} proposed an interesting approach to enforce budget constraints by stopping the CoT or appending "Wait" to prolong it. However, stopping the CoT prematurely  may cause incomplete reasoning and appending "Wait" may risk misalignment with the model’s natural CoT. Alternatively, steering representations may allow for a more coherent way to modulate reasoning length (see Appendix~\ref{sec:steering_examples}) -- by directly manipulating the model’s internal representations, one can more effectively balance the computational cost and performance.


\paragraph{Key insights.} Based on these experiments, we observe that:
\begin{enumerate}[leftmargin=0.5cm]
    \item While steering the model toward longer reasoning (\(\alpha>0\)) does not always guarantee improved performance, steering toward short reasoning (\(\alpha<0\)) consistently degrades accuracy. This suggests that the overly short reasoning with reduced accuracy, as observed in Section~\ref{sec:short_reasoning_prevalence}, is driven by a specific and identifiable pattern in the hidden representations.
    \item Layerwise analysis reveals that the \emph{middle layers} play a key role in regulating reasoning length.
\end{enumerate}
Based on these findings, we hypothesize that certain critical components within the middle layers may contribute to short reasoning. In the next section, we pinpoint these components and perform weight editing to mitigate their effects.

\section{\textit{ThinkEdit}: Mitigate Overly Short Reasoning through Weight Editing}
\label{sec:mitigate_short_reasoning}
Building on the insights from Section~\ref{sec:effects_of_reasoning_length_direction}, in this section, we propose \textbf{\textit{ThinkEdit}}, an effective weight-editing method to mitigate overly short reasoning. We start by analyzing whether specific components within reasoning models significantly contribute to the phenomenon of \emph{short reasoning}. Our focus is on pinpointing particular attention heads, as the attention mechanism plays a crucial role in information propagation across tokens. To explore this, we begin with an overview of the multi-head attention mechanism in Section~\ref{sec:attn_head_overview}, where we define the contribution of individual attention heads. Using this definition, we identify short reasoning heads in Section~\ref{sec:identify_short_reasoning_attn_heads} and remove the short reasoning component from these heads in Section~\ref{sec:editing_short_reasoning_heads}. Finally, in Section~\ref{sec:perf_edited_models}, we evaluate \textbf{\textit{ThinkEdit}}, and show that it effectively mitigates the overly short reasoning issue.

\begin{figure*}[!t]
    \centering
    \includegraphics[width=1.0\textwidth]{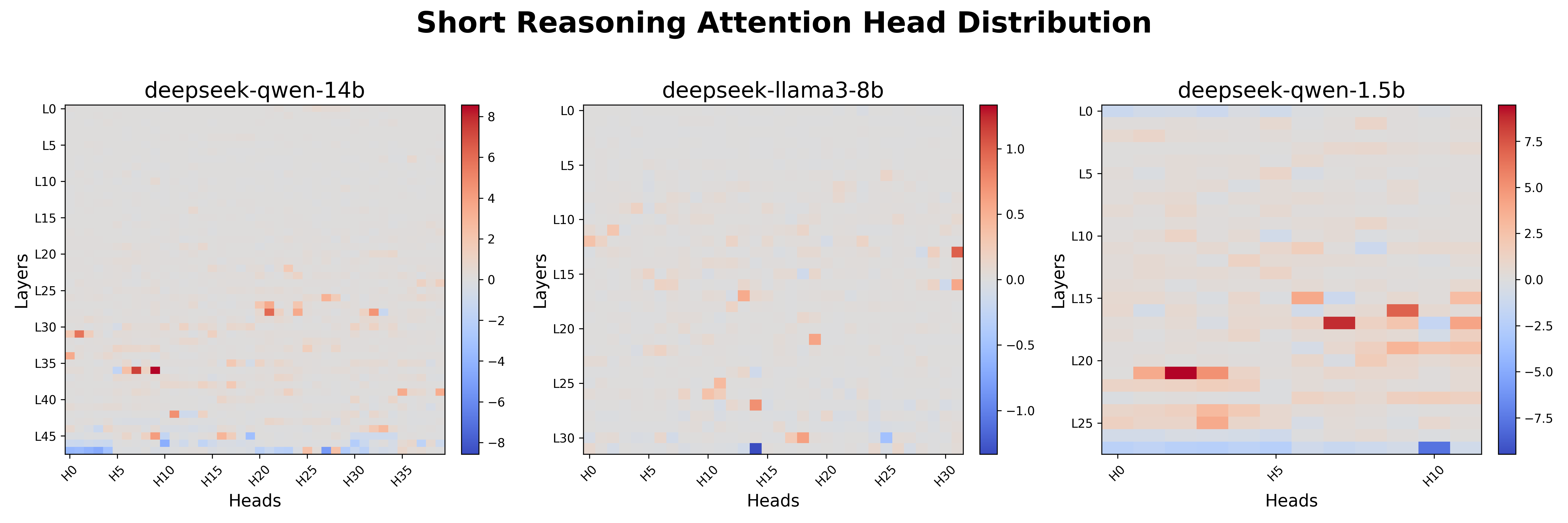}
    \vspace{-20pt}
    \caption{Heatmap illustrating the short reasoning contribution \(\overline{C}^{h}_{\textrm{short}}\) for each attention head \(h\). Heads with higher values (in red) show stronger alignment with short reasoning behavior.}
    \label{fig:short_reasoning_attn_head_distribution}
\end{figure*}

\subsection{Overview of Attention-Head Structure}
\label{sec:attn_head_overview}

A self-attention layer typically includes multiple \emph{attention heads}, each responsible for capturing distinct token-to-token dependencies. Let \( d \) denote the model’s hidden dimension, and \( H \) the number of attention heads. Each head \( h \) operates on a subspace of size \( d_h = \frac{d}{H} \) using the following steps:

\begin{itemize}[leftmargin=1.5em]
    \item \textbf{Q, K, and V Projections.} Given a hidden-state \({r} \in \mathbb{R}^{T \times d}\) (for \(T\) tokens), each head \( h \) computes:
    \[
    Q^h = r W_q^h, \; K^h = r W_k^h, \; V^h = r W_v^h,
    \]
    where \(Q^h, K^h, V^h\!\in\!\mathbb{R}^{T \times d_h}\). Each head \( h \) has its own learnable projection matrices \( {W}_q^h, {W}_k^h, {W}_v^h\!\in\!\mathbb{R}^{d \times d_h} \), which transform the hidden representation \({r}\) into query, key, and value vectors.

    \item \textbf{Self-Attention Computation.} The head outputs an attention-weighted combination of \({V}^h\):
    \vspace{-5pt}
    \[
        {A}^h = \mathrm{softmax}\!\Bigl(\tfrac{{Q}^h ({K}^h)^\top}{\sqrt{d_h}}\Bigr) \, {V}^h
        \;\;\in\; \mathbb{R}^{T \times d_h}.
    \vspace{-5pt}
    \]

    \item \textbf{Output Projection.} Each head’s output \({A}^h\) is merged back into the residual stream via a learned projection matrix \( {W}_o^h \in \mathbb{R}^{d_h \times d} \),
    producing the final \emph{per-head contribution} $C^{h}$:
    \vspace{-5pt}
    \begin{equation}
      C^{h}\coloneq{A}^h {W}_o^h \;\in\; \mathbb{R}^{T \times d}.
      \vspace{-5pt}
    \end{equation}
    The final multi-head attention output is then obtained by summing the contributions from all heads, and this result is added to the residual stream.
\end{itemize} 
The \emph{per-head contribution} $C^{h}$ directly reflects how each attention head modifies the residual stream. This contribution serves as the primary focus of our analysis, as it allows us to pinpoint attention heads that drive \emph{short reasoning} behavior.

\subsection{Identify Short Reasoning Attention Heads}
\label{sec:identify_short_reasoning_attn_heads}

For a response to problem \(i\), let \(\mathcal{T}_i\) be the set of token positions corresponding to the CoT, i.e., the tokens enclosed by \texttt{<think>} and \texttt{</think>} tags. Then, the overall average per-head contribution over all problems in the short reasoning dataset \(\mathcal{D}_{\text{short}}\) is given by $\overline{C}^{h} \in \mathbb{R}^d$:
\vspace{-5pt}
\begin{equation}
\label{e:avg_contribution}
    \overline{C}^{h}=\frac{1}{|\mathcal{D}_{\text{short}}|}\sum_{i\in\mathcal{D}_{\text{short}}} \left( \frac{1}{|\mathcal{T}_i|}\sum_{t\in\mathcal{T}_i} C^{h}(i,t) \right).
    \vspace{-5pt}
\end{equation}
Equation~\ref{e:avg_contribution} first averages the per-head contributions \(C^{h}(i,t)\) over the CoT token positions for each problem $i$ and then averages these values across all problems in \(\mathcal{D}_{\text{short}}\). Recall that the reasoning length direction after an attention layer is defined as
$v^{\text{attn}}_{\ell}=\overline{r}^{\text{attn}}_{\ell,\text{long}}-\overline{r}^{\text{attn}}_{\ell,\text{short}}$ in Equation~\ref{e:reasoning_length_direction}, $v^{\text{attn}}_{\ell} \in \mathbb{R}^d$. To quantify the short reasoning contribution of head \(h\), we project \(\overline{C}^{h}\) onto the negative of the reasoning length direction (i.e., the short reasoning direction). Using the unit vector \(\hat{v}^{\text{attn}}_{\ell} = \frac{v^{\text{attn}}_{\ell}}{\|v^{\text{attn}}_{\ell}\|}\), we define the scalar projection as $\overline{C}^{h}_{\textrm{short}} \in \mathbb{R}$:
\vspace{-5pt}
\begin{equation}
    \overline{C}^{h}_{\textrm{short}} \;=\; \left\langle \overline{C}^{h}, -\hat{v}^{\text{attn}}_{\ell} \right\rangle.
    \vspace{-5pt}
\end{equation}
Here, \(\overline{C}^{h}_{\textrm{short}}\) quantifies the degree to which head \(h\)’s average contribution aligns with the short reasoning direction. Larger values of \(\overline{C}^{h}_{\textrm{short}}\) indicate that the head strongly promotes short reasoning behavior. We visualize \(\overline{C}^{h}_{\textrm{short}}\) for each attention head \(h\) with heatmap in Figure~\ref{fig:short_reasoning_attn_head_distribution}. Only a small subset of heads exhibits notably high alignment with the short reasoning direction, and these heads tend to cluster in the middle layers. This observation aligns with our analysis in section~\ref{sec:effects_of_reasoning_length_direction}, where we found that reasoning length is primarily encoded in the middle layers. Crucially, the sparsity of these "short reasoning heads" suggests that it may be possible to effectively mitigate overly short reasoning behavior with minimal modifications to the model. In the following section, we use these insights to develop a targeted intervention \textbf{\textit{ThinkEdit}} that removes short reasoning components while leaving the vast majority of the model’s parameters unchanged.

\subsection{Editing Short Reasoning Heads}
\label{sec:editing_short_reasoning_heads}

We introduce how \textbf{\textit{ThinkEdit}} effectively removes the short reasoning direction from the output projection matrices of the "short reasoning heads". Specifically, we identify the top 4\% of attention heads with the largest \(\overline{C}^{h}_{\text{short}}\) values (as defined in Section~\ref{sec:identify_short_reasoning_attn_heads}), marking them as short reasoning heads. Let \(W_o^{h_\ell} \in \mathbb{R}^{d_h \times d}\) be the output projection matrix of head \(h\) in layer~\(\ell\), and let \(-\hat{v}^{\text{attn}}_{\ell} \in \mathbb{R}^d\) denote the \emph{short reasoning direction} at layer~\(\ell\). We then update \(W_o^{h_\ell}\) via:
\vspace{-5pt}
\begin{equation}
W_o^{h_\ell} \;\leftarrow\; W_o^{h_\ell} \Bigl(I \;-\; \bigl(-\hat{v}^{\text{attn}}_{\ell}\bigr)\bigl(-\hat{v}^{\text{attn}}_{\ell}\bigr)^{\!\top}\Bigr),
\vspace{-5pt}
\end{equation}
where \(I\) is the \(d \times d\) identity matrix. Intuitively, this operation projects each row of \(W_o^{h_\ell}\) onto the subspace orthogonal to \(-\hat{v}^{\text{attn}}_{\ell}\), thereby removes the short reasoning component from the head’s output contribution. Unlike the approach in Section~\ref{sec:effects_of_reasoning_length_direction}, which adds a fixed direction to activations regardless of the input, \textbf{\textit{ThinkEdit}} modifies the weights of selected attention heads. This makes the adjustment input-dependent, allowing more precise control over reasoning length while preserving the model’s overall behavior.

\begin{table*}[t]
\centering
\scriptsize
\begin{tabular*}{\linewidth}{@{\extracolsep{\fill}} l l c c c c c}
\toprule
\textbf{Model} & & \textbf{GSM8K} & \textbf{MMLU Elem. Math} & \textbf{MATH-Level1} & \textbf{MATH-Level5} & \textbf{MATH-500} \\
\midrule
\multirow{2}{*}{\textbf{deepseek-qwen-14B}} 
 & Original & $90.80 \pm 0.36$ & $95.08 \pm 0.65$ & $96.32 \pm 0.35$ & $90.25 \pm 0.72$ & $91.48 \pm 0.55$ \\
 & ThinkEdit (4\%) & $\mathbf{93.78 \pm 0.50}$ & $\mathbf{96.56 \pm 0.84}$ & $\mathbf{96.36 \pm 0.52}$ & $\mathbf{91.03 \pm 0.44}$ & $\mathbf{91.92 \pm 0.63}$ \\
\midrule
\multirow{2}{*}{\textbf{deepseek-llama3-8B}}
 & Original & $82.26 \pm 0.91$ & $96.01 \pm 0.62$ & $93.46 \pm 0.84$ & $85.49 \pm 0.83$ & $87.26 \pm 1.16$ \\
 & ThinkEdit (4\%) & $\mathbf{89.44 \pm 0.55}$ & $\mathbf{96.19 \pm 0.73}$ & $\mathbf{94.44 \pm 0.31}$ & $\mathbf{86.49 \pm 0.54}$ & $\mathbf{88.06 \pm 1.09}$ \\
\midrule
\multirow{2}{*}{\textbf{deepseek-qwen-1.5B}}
 & Original & $79.15 \pm 1.08$ & $68.52 \pm 1.56$ & $93.00 \pm 0.33$ & $\mathbf{75.48 \pm 0.90}$ & $82.22 \pm 1.29$ \\
 & ThinkEdit (4\%) & $\mathbf{84.56 \pm 0.79}$ & $\mathbf{90.66 \pm 0.97}$ & $\mathbf{93.66 \pm 0.62}$ & $75.05 \pm 0.82$ & $\mathbf{82.24 \pm 0.89}$ \\
\bottomrule
\end{tabular*}
\vspace{-10pt}
\caption{Overall accuracy (\%) of each model before and after applying ThinkEdit.}
\label{tab:overall_accuracy}
\end{table*}

\begin{table*}[t]
\centering
\setlength{\tabcolsep}{4pt} 
\scriptsize
\begin{tabular*}{\linewidth}{@{\extracolsep{\fill}} l l c c c c c}
\toprule
\textbf{Model} & & \textbf{GSM8K} & \textbf{MMLU Elem. Math} & \textbf{MATH-Level1} & \textbf{MATH-Level5} & \textbf{MATH-500} \\
\midrule
\multirow{2}{*}{\textbf{deepseek-qwen-14b}} 
 & Original & \textbf{96.31} / 95.65 / 92.93 & 93.89 / \textbf{96.22} / 95.60 & \textbf{99.52} / 99.30 / 97.70 & 89.39 / 94.32 / 96.25 & 86.40 / 91.40 / 93.50 \\
 & ThinkEdit (4\%) & \textbf{96.31} / \textbf{96.18} / \textbf{96.77} & \textbf{97.78} / 95.14 / \textbf{96.53} & \textbf{99.52} / \textbf{99.53} / \textbf{98.62} & \textbf{96.67} / \textbf{97.88} / \textbf{98.11} & \textbf{91.20} / \textbf{93.20} / \textbf{95.00} \\
\midrule
\multirow{2}{*}{\textbf{deepseek-llama3-8b}} 
 & Original & 88.92 / 87.18 / 85.82 & 97.22 / 96.49 / 96.80 & 97.14 / 94.88 / 94.83 & 78.64 / 88.79 / 93.41 & 82.00 / 81.40 / 88.30 \\
 & ThinkEdit (4\%) & \textbf{96.31} / \textbf{95.50} / \textbf{94.68} & \textbf{97.78} / \textbf{97.57} / \textbf{97.60} & \textbf{99.05} / \textbf{99.07} / \textbf{97.82} & \textbf{95.76} / \textbf{97.42} / \textbf{97.46} & \textbf{95.60} / \textbf{93.80} / \textbf{95.40} \\
\midrule
\multirow{2}{*}{\textbf{deepseek-qwen-1.5b}} 
 & Original & 88.46 / 87.48 / 85.02 & 62.78 / 62.16 / 60.53 & \textbf{97.62} / 95.12 / 93.91 & 91.52 / 95.00 / 95.72 & 82.40 / 89.80 / 93.40 \\
 & ThinkEdit (4\%) & \textbf{92.62} / \textbf{92.90} / \textbf{92.32} & \textbf{87.78} / \textbf{88.11} / \textbf{88.67} & 95.71 / \textbf{95.58} / \textbf{96.44} & \textbf{95.15} / \textbf{96.59} / \textbf{97.27} & \textbf{90.80} / \textbf{92.00} / \textbf{94.20} \\
\bottomrule
\end{tabular*}
\vspace{-10pt}
\caption{Accuracy (\%) of the top 5\% / 10\% / 20\% shortest reasoning responses.}
\label{tab:short_reasoning_acc}
\end{table*}

\subsection{Performance of Reasoning Models after \textbf{\textit{ThinkEdit}}}
\label{sec:perf_edited_models}

\paragraph{Experimental Setup.}
We evaluate the reasoning models after applying \textbf{\textit{ThinkEdit}} on four mathematical reasoning benchmarks:

\begin{itemize}[leftmargin=1.5em] 
    \item \textbf{GSM8K} \cite{gsm8k}: A test set of 1,319 grade-school-level math word problems. \item \textbf{MMLU Elementary Math} \cite{mmlu}: A subset of 378 elementary school math questions from the MMLU benchmark.
    \item \textbf{MATH-Level1}: A collection of 437 easy (Level 1) problems drawn from the MATH dataset \cite{math}.
    \item \textbf{MATH-Level5}: The most challenging subset of the MATH dataset with 1,324 problems.
    \item \textbf{MATH-500} \cite{math500}: A curated set of 500 high-quality math problems designed to assess advanced mathematical reasoning.
\end{itemize}
For all datasets, we set a maximum CoT length of 16,384 tokens. If this limit is reached, the model is prompted to immediately finalize its answer. To mitigate randomness, each dataset is evaluated over 10 independent runs, and the mean accuracy is reported. We do not include the phrase "Please reason step by step" in any prompt, aiming to assess the model’s inherent reasoning capabilities.

\paragraph{Overall Accuracy.}
Table~\ref{tab:overall_accuracy} reports the overall accuracy (in \%) before and after applying \textbf{\textit{ThinkEdit}}. Across all math benchmarks, we observe consistent improvements in accuracy. Notably, the \texttt{deepseek-distill-qwen-1.5B} model shows a substantial gain on the MMLU Elementary Math subset. Manual inspection reveals that the unedited model occasionally ignores the multiple-choice format, leading to wrong answers. In contrast, the edited model adheres to the instructions more reliably. This suggests that \textbf{\textit{ThinkEdit}} may not only enhance reasoning quality but also improve instruction-following behavior. On the more challenging MATH-Level5 and MATH-500 datasets, the accuracy gains are more modest but still positive, suggesting that while editing short-reasoning heads has a stronger impact on simpler problems, it might still provide meaningful improvements even for harder tasks that require longer and more complex reasoning chains.

\paragraph{Accuracy Under Short Reasoning.}
Table~\ref{tab:short_reasoning_acc} shows the average accuracy for the top 5\%, 10\%, and 20\% of responses with the shortest reasoning traces. After applying \textbf{\textit{ThinkEdit}}, we observe substantial accuracy improvements in these short-reasoning cases across most benchmarks. Interestingly, even for the challenging MATH-Level5 and MATH-500 datasets, short-reasoning accuracy improves noticeably. This suggests that \textbf{\textit{ThinkEdit}} can effectively improve the reasoning quality when the models generate short CoT.

\paragraph{Reasoning Length of the Shortest Responses.} We analyze how \textbf{\textit{ThinkEdit}} affects reasoning length in Appendix~\ref{sec:thinkedit_reasoning_length}. It modestly increases the length of the shortest responses (Table~\ref{tab:short_reasoning_len}), helping to address overly brief reasoning. However, as shown in Table~\ref{tab:overall_reasoning_length}, the overall reasoning length remains largely stable across datasets, with a net change of -0.27\% across all models and datasets.

\vspace{0.9\baselineskip}

\noindent In summary, \textbf{\textit{ThinkEdit}} markedly improves model performance on short-reasoning instances and yields a substantial overall accuracy gain. We also explore different editing percentages and compare our approach to simply appending “Wait” to prompt longer reasoning; detailed results are provided in Appendix~\ref{sec:full_results}. Additionally, results for \texttt{deepseek-distill-qwen-32B} are reported in Appendix~\ref{sec:32b_results}. To test the generality of our method, we further evaluate on non-math domains in Appendix~\ref{app:non_math_mmlu}. Beyond demonstrating accuracy gains, Appendix~\ref{app:behavioral_analysis} examines how \textbf{\textit{ThinkEdit}} shapes reasoning behaviors, revealing more explicit and structured chains of thought. Finally, We provide several concrete examples illustrating how \textbf{\textit{ThinkEdit}} enhances reasoning quality in Appendix~\ref{sec:examples_editing}.

\section{Related Works}
\label{sec: related works}
\paragraph{Reasoning Models.}
Recent advances in reasoning models have significantly improved the problem-solving abilities of LLMs in domains such as mathematics, coding, and science. OpenAI's \texttt{o1}~\cite{o1} represents a major shift toward deliberate reasoning by employing reinforcement learning (RL) to refine its strategies. By generating explicit "Thinking" steps before producing answers, \texttt{o1} achieves strong performance on complex tasks. As a more cost-efficient alternative, DeepSeek-\texttt{r1}~\cite{r1} demonstrates that pure RL can also effectively enhance reasoning. It introduces Group Relative Policy Optimization (GRPO)~\cite{grpo}, a novel method that eliminates the need for a separate reward model, enabling more efficient RL training.
 


\paragraph{Controllable Text Generation.}
Controllable text generation has been explored across various domains \cite{control_survey}, with methods generally classified into training-time and inference-time control. These approaches aim to steer LLMs toward generating text with specific attributes while preserving fluency and coherence. Training-time control is achieved through fine-tuning \cite{auxiliary_tuning,flan} or reinforcement learning \cite{rlhf,dpo}, leveraging diverse datasets to shape model behavior. Inference-time control encompasses prompt engineering \cite{autoprompt,prefixtuning}, representation engineering \cite{steering_vector,representation_engineering,style_vector,neuroneval}, interpretable neuron intervention through concept bottleneck models \cite{cbllm,cbllm_classification}, and decoding-time interventions \cite{pplm}, allowing flexible and efficient control without retraining. 

In this work, we focus on the representation engineering paradigm to investigate how reasoning length is embedded within model representations. Specifically, we introduce a linear "reasoning-length direction" in the representation space to examine its impact on reasoning capabilities.

\paragraph{Attention heads and MLP neurons intervention.} A growing body of research explores the role of attention heads and neurons within the Multi-Layer Perceptron (MLP) layers in shaping language model behavior. Studies such as \cite{attnsafety, safetyneurons1, safetyneurons2} examine how safety mechanisms are embedded in well-aligned models to defend against harmful prompts and jailbreak attacks \cite{gcg, autodan, advllm}. Findings indicate that a small subset of attention heads and MLP neurons play a critical role in safety alignments. 
Similarly, research on hallucination mitigation has investigated the contributions of attention heads and MLP neurons. \cite{novo} demonstrates that filtering out unreliable attention heads can reduce erroneous generations, while \cite{MechnisticHallucination} finds that activating subject-knowledge neurons helps maintain factual consistency. In~\cite{neuronadjust}, the authors design efficient and training-free machine skill unlearning techniques for LLMs through intervention and abstention.

In our work, we investigate how attention heads relate to reasoning processes, examining their impact on the reasoning length and quality.

\section{Conclusion}
In this work, we first identified overly short reasoning as a common failure mode in Deepseek-distilled reasoning models. To understand how reasoning length is controlled, we analyzed the model’s hidden representations and uncovered a latent direction linked to reasoning length. Building on this, we pinpointed 4\% of attention heads that drive short reasoning, and propose \textbf{\textit{ThinkEdit}} to mitigate the issue, leading to significant accuracy gains for short reasoning outputs (+6.39\%), along with an overall improvement (+3.34\%) across multiple math benchmarks.


\section*{Limitations}
A limitation of our work is that \textbf{\textit{ThinkEdit}} primarily improves model performance by addressing cases of overly short reasoning. For reasoning models that already tend to produce sufficiently long or verbose outputs, the benefits of \textbf{\textit{ThinkEdit}} may be limited. Nonetheless, our study provides valuable insights by highlighting the often-overlooked issue of overly brief reasoning and examining how reasoning length is represented within the model's hidden states. This opens an important research direction for advancing the interpretability of reasoning models by linking internal representations to observable reasoning behaviors.

\section*{Acknowledgement}

This research was supported by grants from NVIDIA and utilized NVIDIA’s A100 GPUs on Saturn Cloud. The authors are partially supported by Hellman Fellowship, Intel Rising Star Faculty Award, and National Science Foundation under Grant No. 2313105, 2430539.

\bibliography{custom}

\clearpage

\appendix
\onecolumn
\addcontentsline{toc}{section}{Appendix} 
\part{} 
\parttoc 

\section{Appendix}

\subsection{Gobal Steering with the MLP-based Direction \(v_{\ell}^{\text{mlp}}\)}
\label{sec:mlp_steering}
\begin{figure}[H]
    \centering
    \includegraphics[width=0.8\textwidth]{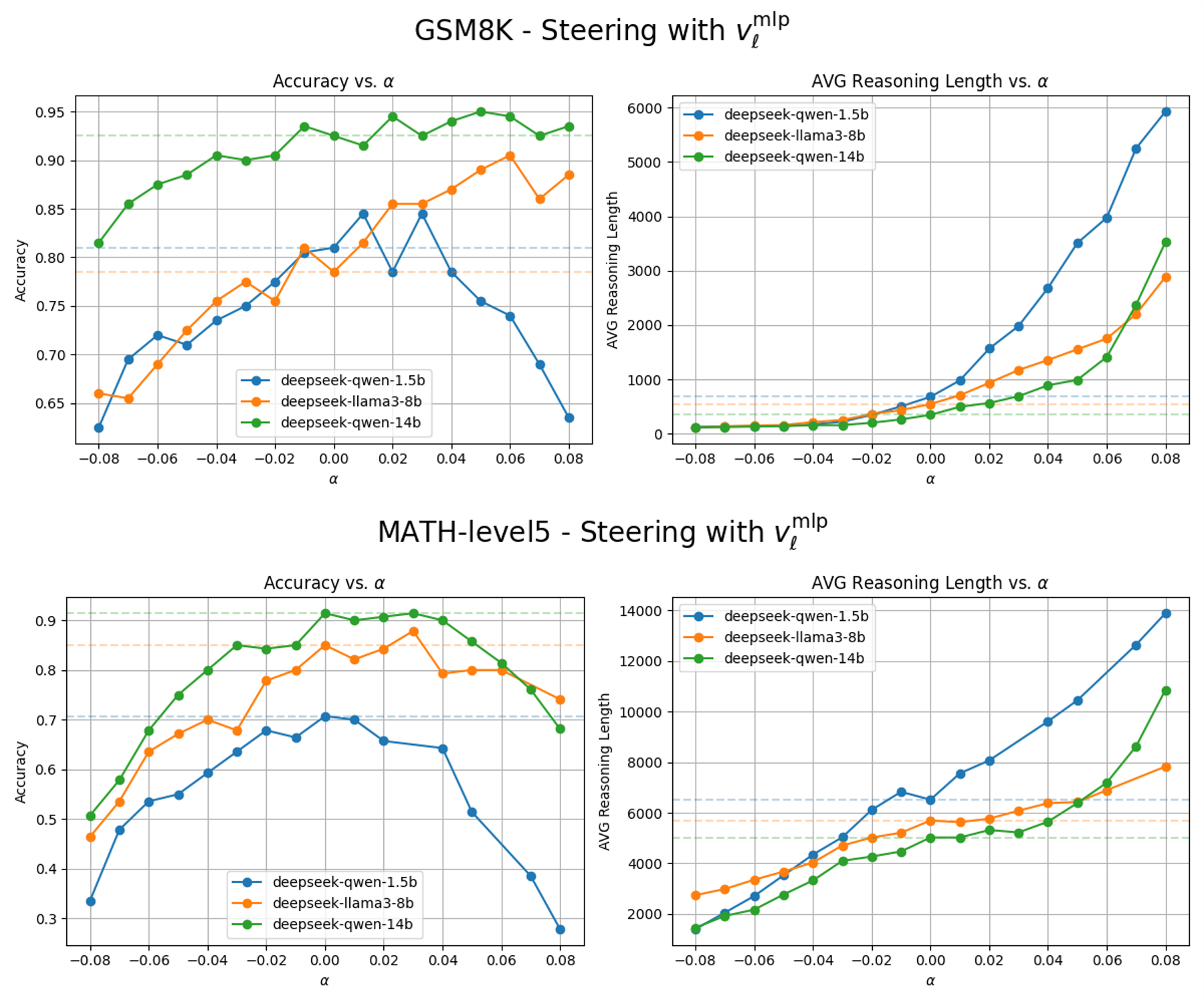}
    \vspace{-10pt}
    \caption{
        Global steering with the reasoning length direction extracted from MLPs. The trend is similar as steering with attention-based directions.
    }
    \label{fig:steering_mlp}
\end{figure}

Figure~\ref{fig:steering_mlp} replicates the global steering analysis using the MLP-based direction \(v_{\ell}^{\text{mlp}}\). The observed trends closely mirror those from attention-based steering: increasing \(\alpha\) extends reasoning length across both datasets, and the effect on accuracy is model- and dataset-dependent. On GSM8K, larger models benefit from longer reasoning, while smaller models degrade. On MATH-Level5, overly long reasoning may hurt performance, despite consistent control over CoT length. These results show that both attention and MLP directions capture similar reasoning-length attributes.

\clearpage

\subsection{Layerwise Analysis of Steering along Reasoning Length Direction}
\label{sec:layerwise_steering}

\begin{figure}[H]
    \centering
    \includegraphics[width=0.95\textwidth]{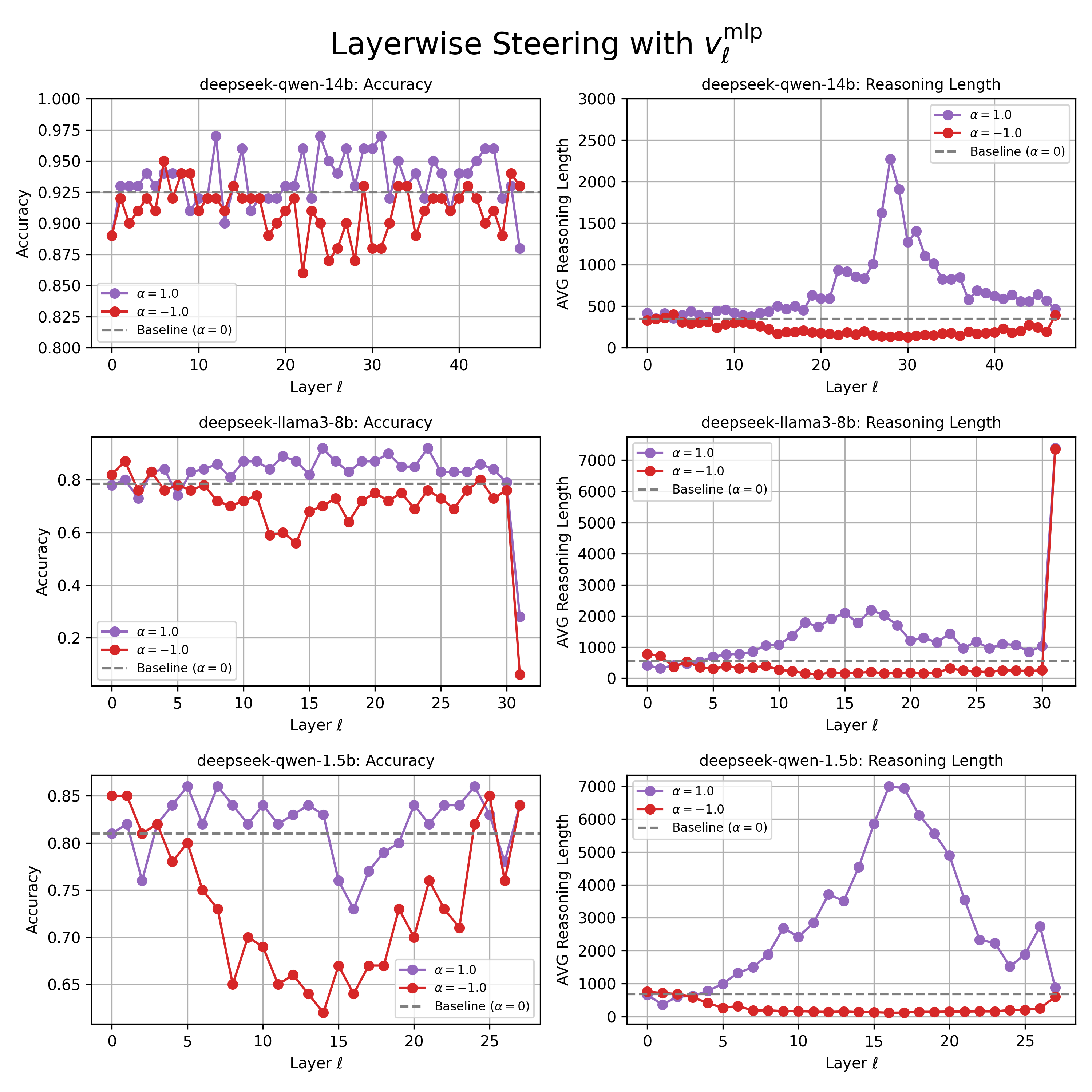}
    \vspace{-10pt}
    \caption{Layerwise steering on GSM8K with $v^{mlp}_{\ell}$. We apply \(\alpha=\pm1\) to one layer at a time, revealing that the middle layers wield the strongest control over reasoning length and accuracy.}
    \label{fig:gsm8k_layerwise_steering}
\end{figure}

To identify which layers are most influenced by the reasoning-length direction, we perform a \emph{layerwise} experiment on the GSM8K dataset (Figure~\ref{fig:gsm8k_layerwise_steering}). Specifically, we use $v^{\text{mlp}}_{\ell}$ to steer each MLP layer \emph{separately}, applying $\alpha = \pm 1$ at a single layer $\ell$. Notably, the \emph{middle layers} elicit the largest fluctuations, suggesting they encode key representations for controlling reasoning length.

\clearpage

\subsection{The Impact of ThinkEdit on Reasoning Length}
\label{sec:thinkedit_reasoning_length}

Table~\ref{tab:short_reasoning_len} reports the average reasoning length among the top 5\%, 10\%, and 20\% shortest responses. We observe that \textbf{\textit{ThinkEdit}} consistently increases the reasoning length in these short-answer scenarios, suggesting that the intervention discourages excessively short reasoning, and therefore leads to higher accuracy as shown in Table~\ref{tab:short_reasoning_acc}. Interestingly, Table~\ref{tab:overall_reasoning_length} shows that the average reasoning length remains similar between the original and \textbf{\textit{ThinkEdit}} models. To summarize these trends, we compute the average change in reasoning length across all datasets: +2.94\% for \texttt{deepseek-qwen-14b}, +3.53\% for \texttt{deepseek-llama3-8b}, and -5.73\% for \texttt{deepseek-qwen-1.5b}, resulting in an overall average change of -0.27\%. These results suggest that \textbf{\textit{ThinkEdit}} selectively increases reasoning length for short responses without significantly altering overall response length.

\setlength{\tabcolsep}{4pt}
\begin{table}[H]
\centering
\scriptsize
\begin{tabular*}{\linewidth}{@{\extracolsep{\fill}} l l c c c c c}
\toprule
\textbf{Model} & & \textbf{GSM8K} & \textbf{MMLU Elem. Math} & \textbf{MATH-Level1} & \textbf{MATH-Level5} & \textbf{MATH-500} \\
\midrule
\multirow{2}{*}{\textbf{deepseek-qwen-14B}} 
 & Original & 76.6 / 86.5 / 99.1 & 65.8 / 72.2 / 80.6 & 93.7 / 114.3 / 188.6 & 628.8 / 858.4 / 1125.9 & 198.7 / 434.3 / 697.0 \\
 & ThinkEdit (4\%) & \textbf{101.7} / \textbf{113.6} / \textbf{131.0} & \textbf{82.7} / \textbf{91.8} / \textbf{105.6} & \textbf{146.7} / \textbf{188.6} / \textbf{346.0} & \textbf{745.5} / \textbf{926.6} / \textbf{1163.7} & \textbf{361.3} / \textbf{559.3} / \textbf{764.6} \\
\midrule
\multirow{2}{*}{\textbf{deepseek-llama3-8B}}
 & Original & 73.0 / 83.1 / 96.6 & 371.0 / 438.1 / 518.2 & 80.3 / 97.2 / 130.3 & 617.9 / 854.9 / 1126.5 & 159.5 / 357.5 / 644.5 \\
 & ThinkEdit (4\%) & \textbf{110.3} / \textbf{131.8} / \textbf{164.6} & \textbf{398.5} / \textbf{462.4} / \textbf{541.8} & \textbf{176.3} / \textbf{221.3} / \textbf{336.0} & \textbf{806.1} / \textbf{963.3} / \textbf{1185.1} & \textbf{372.5} / \textbf{553.2} / \textbf{772.9} \\
\midrule
\multirow{2}{*}{\textbf{deepseek-qwen-1.5B}}
 & Original & 78.8 / 89.4 / 103.0 & 61.6 / 68.5 / 77.6 & 88.8 / 110.3 / 219.7 & 804.6 / \textbf{1017.9} / \textbf{1314.0} & 249.7 / 506.5 / 760.7 \\
 & ThinkEdit (4\%) & \textbf{103.3} / \textbf{118.9} / \textbf{144.8} & \textbf{80.6} / \textbf{92.5} / \textbf{112.9} & \textbf{172.7} / \textbf{336.9} / \textbf{543.6} & \textbf{853.0} / 1003.5 / 1221.9 & \textbf{530.8} / \textbf{676.0} / \textbf{837.4} \\
\bottomrule
\end{tabular*}
\vspace{-10pt}
\caption{Average reasoning length for the top 5\% / 10\% / 20\% shortest responses (in tokens).}
\label{tab:short_reasoning_len}
\end{table}

\setlength{\tabcolsep}{4pt}
\begin{table}[H]
\centering
\scriptsize
\begin{tabular*}{\linewidth}{@{\extracolsep{\fill}} l l c c c c c}
\toprule
\textbf{Model} & & \textbf{GSM8K} & \textbf{MMLU Elem. Math} & \textbf{MATH-Level1} & \textbf{MATH-Level5} & \textbf{MATH-500} \\
\midrule
\multirow{2}{*}{\textbf{deepseek-qwen-14B}} 
 & Original    & $354.5 \pm 684.4$               & $184.9 \pm 175.3$               & $1600.5 \pm 1885.2$              & $\mathbf{4306.2 \pm 3816.1}$     & $\mathbf{3096.8 \pm 3308.0}$     \\
 & ThinkEdit (4\%)  & $\mathbf{538.2 \pm 829.6}$      & $\mathbf{291.4 \pm 607.5}$      & $\mathbf{1670.4 \pm 1951.2}$     & $4243.7 \pm 3814.0$              & $3079.7 \pm 3276.6$              \\
\midrule
\multirow{2}{*}{\textbf{deepseek-llama3-8B}}
 & Original    & $597.3 \pm 1109.0$             & $1486.6 \pm 2036.7$             & $1646.6 \pm 2275.0$             & $\mathbf{4789.1 \pm 4315.4}$     & $3507.6 \pm 3917.5$              \\
 & ThinkEdit (4\%)  & $\mathbf{927.7 \pm 1486.3}$     & $\mathbf{1517.9 \pm 2041.5}$    & $\mathbf{1723.7 \pm 2152.3}$    & $4773.5 \pm 4327.4$              & $\mathbf{3509.5 \pm 3842.9}$     \\
\midrule
\multirow{2}{*}{\textbf{deepseek-qwen-1.5B}}
 & Original    & $768.1 \pm 1837.2$             & $517.0 \pm 1502.8$              & $\mathbf{2080.9 \pm 2740.5}$    & $\mathbf{6360.0 \pm 5336.4}$     & $\mathbf{4260.3 \pm 4668.2}$     \\
 & ThinkEdit (4\%)  & $\mathbf{1126.6 \pm 2018.0}$    & $\mathbf{768.9 \pm 1651.4}$     & $1946.3 \pm 2438.4$             & $5522.4 \pm 5036.9$              & $3821.1 \pm 4384.9$              \\
\bottomrule
\end{tabular*}
\vspace{-10pt}
\caption{Overall reasoning length (in tokens) before and after applying ThinkEdit (4\% edit rate).}
\label{tab:overall_reasoning_length}
\end{table}

\clearpage

\subsection{ThinkEdit with Varying Editing Rates vs. the "Wait" Appending Baseline}
\label{sec:full_results}

We conduct a comprehensive evaluation of \textbf{\textit{ThinkEdit}} with different editing rates and compare it against a simple baseline that appends the word "Wait" to reasoning sequences shorter than 500 tokens. This baseline is intended to prompt the model to continue thinking before answering when the reasoning is too short. The methods compared are:

\begin{itemize}
\item \textbf{\textit{ThinkEdit (8\%)}}: Edits 8\% of total attention heads.
\item \textbf{\textit{ThinkEdit (4\%)}}: Edits 4\% of total attention heads.
\item \textbf{\textit{ThinkEdit (2\%)}}: Edits 2\% of total attention heads.
\item \textbf{Append "Wait"}: Appends "Wait" to reasoning with fewer than 500 tokens to artificially extend reasoning length.
\item \textbf{Original}: The unmodified model output.
\end{itemize}
As shown in Table~\ref{tab:overall_acc_detailed}, higher editing rates in \textbf{\textit{ThinkEdit}} consistently improve performance on GSM8K and MMLU Elementary Math. However, for the MATH-series datasets, moderate editing rates yield better results than the most aggressive edits. The "Append Wait" baseline yields only marginal gains across all datasets, indicating that simply forcing the model to produce longer reasoning does not necessarily improve accuracy. A closer look at the short reasoning cases in Table~\ref{tab:short_acc_detailed} shows that all versions of \textbf{\textit{ThinkEdit}} substantially outperform the "Append Wait" baseline. This further supports the observation that longer reasoning alone is insufficient without proper internal adjustment of the model.

In terms of reasoning length (Tables~\ref{tab:overall_len_detailed} and~\ref{tab:short_len_detailed}), the "Append Wait" method generally leads to longer outputs than \textbf{\textit{ThinkEdit (2\%)}}, confirming that it effectively increases response length. However, despite this, it fails to match the performance of \textbf{\textit{ThinkEdit}}, highlighting that \textbf{\textit{ThinkEdit}} is a more effective strategy addressing the accuracy drops of overly short reasoning.

\begin{table}[H]
\centering
\scriptsize
\begin{tabular*}{\linewidth}{@{\extracolsep{\fill}} l l c c c c c}
\toprule
\textbf{Model} & & \textbf{GSM8K} & \textbf{MMLU Elem.\ Math} & \textbf{MATH-Level1} & \textbf{MATH-Level5} & \textbf{MATH-500} \\
\midrule
\multirow{5}{*}{\textbf{deepseek-qwen-14B}}
 & ThinkEdit (8\%)  & $\mathbf{94.30\pm0.28}$ & $\mathbf{96.93\pm0.50}$ & $96.09\pm0.35$ & $90.92\pm0.41$ & $91.26\pm0.52$ \\
 & ThinkEdit (4\%)  & $93.78\pm0.50$ & $96.56\pm0.84$ & $96.36\pm0.52$ & $91.03\pm0.44$ & $\mathbf{91.92\pm0.63}$ \\
 & ThinkEdit (2\%)  & $93.50\pm0.31$ & $96.53\pm0.54$ & $96.50\pm0.46$ & $\mathbf{91.15\pm0.59}$ & $91.78\pm0.58$ \\
 & Append "Wait"             & $91.30\pm0.55$ & $95.37\pm0.70$ & $\mathbf{96.52\pm0.55}$ & $90.60\pm0.41$ & $91.22\pm0.57$ \\
 & Original         & $90.80\pm0.36$ & $95.08\pm0.65$ & $96.32\pm0.35$ & $90.25\pm0.72$ & $91.48\pm0.55$ \\
\midrule
\multirow{5}{*}{\textbf{deepseek-llama3-8B}}
 & ThinkEdit (8\%)  & $\mathbf{90.18\pm0.60}$ & $96.11\pm0.67$ & $94.39\pm0.61$ & $86.13\pm0.46$ & $87.64\pm0.88$ \\
 & ThinkEdit (4\%)  & $89.44\pm0.55$ & $\mathbf{96.19\pm0.73}$ & $\mathbf{94.44\pm0.31}$ & $\mathbf{86.49\pm0.54}$ & $\mathbf{88.06\pm1.09}$ \\
 & ThinkEdit (2\%)  & $88.97\pm0.78$ & $96.08\pm0.86$ & $94.12\pm0.47$ & $85.91\pm0.48$ & $87.60\pm0.81$ \\
 & Append "Wait"             & $84.28\pm0.64$ & $95.93\pm0.70$ & $93.96\pm0.55$ & $85.33\pm0.79$ & $87.66\pm1.26$ \\
 & Original         & $82.26\pm0.91$ & $96.01\pm0.62$ & $93.46\pm0.84$ & $85.49\pm0.83$ & $87.26\pm1.16$ \\
\midrule
\multirow{5}{*}{\textbf{deepseek-qwen-1.5B}}
 & ThinkEdit (8\%)  & $\mathbf{84.81\pm0.69}$ & $\mathbf{92.38\pm1.04}$ & $\mathbf{94.00\pm0.75}$ & $75.32\pm1.11$ & $82.56\pm1.21$ \\
 & ThinkEdit (4\%)  & $84.56\pm0.79$ & $90.66\pm0.97$ & $93.66\pm0.62$ & $75.05\pm0.82$ & $82.24\pm0.89$ \\
 & ThinkEdit (2\%)  & $83.34\pm0.79$ & $86.24\pm1.12$ & $93.89\pm0.76$ & $74.94\pm0.85$ & $82.74\pm0.77$ \\
 & Append "Wait"             & $79.81\pm0.77$ & $76.64\pm1.18$ & $93.34\pm0.86$ & $75.06\pm0.72$ & $\mathbf{82.98\pm1.00}$ \\
 & Original         & $79.15\pm1.08$ & $68.52\pm1.56$ & $93.00\pm0.33$ & $\mathbf{75.48\pm0.90}$ & $82.22\pm1.29$ \\
\bottomrule
\end{tabular*}
\vspace{-10pt}
\caption{Overall accuracy (\%) of ThinkEdit with different editing rates.}
\label{tab:overall_acc_detailed}
\end{table}
\setlength{\tabcolsep}{4pt}
\begin{table}[H]
\centering
\scriptsize
\begin{tabular*}{\linewidth}{@{\extracolsep{\fill}} l l c c c c c}
\toprule
\textbf{Model} & & \textbf{GSM8K} & \textbf{MMLU Elem.\ Math} & \textbf{MATH-Level1} & \textbf{MATH-Level5} & \textbf{MATH-500} \\
\midrule
\multirow{5}{*}{\textbf{deepseek-qwen-14B}}
 & ThinkEdit (8\%)  & $\mathbf{598.1\pm1011.8}$ & $\mathbf{336.6\pm550.3}$  & $1586.1\pm1827.4$ & $4150.5\pm3819.1$ & $3009.5\pm3336.7$ \\
 & ThinkEdit (4\%)  & $538.2\pm829.6$           & $291.4\pm607.5$           & $\mathbf{1670.4\pm1951.2}$ & $4243.7\pm3814.0$ & $3079.7\pm3276.6$ \\
 & ThinkEdit (2\%)  & $479.8\pm968.5$           & $285.1\pm756.8$           & $1645.4\pm1946.6$ & $\mathbf{4327.2\pm3863.4}$ & $\mathbf{3138.3\pm3372.8}$ \\
 & Append "Wait"             & $447.3\pm652.6$           & $273.0\pm215.8$           & $1595.8\pm1810.5$ & $4265.9\pm3749.0$ & $3071.5\pm3275.6$ \\
 & Original         & $354.5\pm684.4$           & $184.9\pm175.3$           & $1600.5\pm1885.2$ & $4306.2\pm3816.1$ & $3096.8\pm3308.0$ \\
\midrule
\multirow{5}{*}{\textbf{deepseek-llama3-8B}}
 & ThinkEdit (8\%)  & $\mathbf{971.8\pm1447.7}$          & $1488.3\pm1979.5$         & $1692.8\pm2200.5$ & $4642.1\pm4253.3$ & $3463.3\pm3800.1$ \\
 & ThinkEdit (4\%)  & $927.7\pm1486.3$          & $1517.9\pm2041.5$         & $1723.7\pm2152.3$ & $4773.5\pm4327.4$ & $\mathbf{3509.5\pm3842.9}$ \\
 & ThinkEdit (2\%)  & $849.7\pm1454.8$          & $\mathbf{1520.1\pm2103.0}$ & $\mathbf{1765.7\pm2315.1}$ & $\mathbf{4825.2\pm4383.4}$ & $3503.8\pm3838.4$ \\
 & Append "Wait"             & $670.2\pm1073.0$          & $1514.4\pm2009.1$         & $1639.9\pm2134.8$ & $4795.3\pm4296.2$ & $3502.5\pm3859.1$ \\
 & Original         & $597.3\pm1109.0$          & $1486.6\pm2036.7$         & $1646.6\pm2275.0$ & $4789.1\pm4315.4$ & $3507.6\pm3917.5$ \\
\midrule
\multirow{5}{*}{\textbf{deepseek-qwen-1.5B}}
 & ThinkEdit (8\%)  & $\mathbf{1166.2\pm1986.4}$ & $\mathbf{890.7\pm1851.7}$  & $1912.8\pm2287.6$ & $5567.4\pm5083.4$ & $3772.6\pm4296.0$ \\
 & ThinkEdit (4\%)  & $1126.6\pm2018.0$         & $768.9\pm1651.4$          & $1946.3\pm2438.4$ & $5522.4\pm5036.9$ & $3821.1\pm4384.9$ \\
 & ThinkEdit (2\%)  & $912.7\pm1835.3$          & $701.0\pm1748.9$          & $1918.0\pm2420.6$ & $5641.9\pm5101.5$ & $3880.3\pm4402.4$ \\
 & Append "Wait"             & $847.1\pm1835.7$          & $660.1\pm1823.7$          & $\mathbf{2163.7\pm2847.0}$ & $\mathbf{6363.9\pm5352.9}$ & $\mathbf{4287.1\pm4710.3}$ \\
 & Original         & $768.1\pm1837.2$          & $517.0\pm1502.8$          & $2080.9\pm2740.5$ & $6360.0\pm5336.4$ & $4260.3\pm4668.2$ \\
\bottomrule
\end{tabular*}
\vspace{-10pt}
\caption{Overall reasoning length (in tokens) of ThinkEdit with different editing rates.}
\label{tab:overall_len_detailed}
\end{table}

\begin{table}[H]
\centering
\setlength{\tabcolsep}{4pt}
\scriptsize
\begin{tabular*}{\linewidth}{@{\extracolsep{\fill}} l l c c c c c}
\toprule
\textbf{Model} & & \textbf{GSM8K} & \textbf{MMLU Elem.\ Math} & \textbf{MATH-Level1} & \textbf{MATH-Level5} & \textbf{MATH-500}\\
\midrule
\multirow{5}{*}{\textbf{deepseek-qwen-14B}}
 & ThinkEdit (8\%)  & 96.46 / \textbf{97.02} / \textbf{97.38}
                   & 97.22 / 95.95 / 95.73
                   & 98.57 / 97.91 / 97.47
                   & \textbf{98.48} / \textbf{98.56} / \textbf{98.22}
                   & \textbf{91.60} / 93.00 / 94.60 \\
 & ThinkEdit (4\%)  & 96.31 / 96.18 / 96.77
                   & \textbf{97.78} / 95.14 / 96.53
                   & 99.52 / 99.53 / 98.62
                   & 96.67 / 97.88 / 98.11
                   & 91.20 / \textbf{93.20} / \textbf{95.00} \\
 & ThinkEdit (2\%)  & \textbf{96.62} / 96.03 / 96.12
                   & 96.11 / 96.22 / 96.27
                   & \textbf{100.00} / \textbf{99.77} / \textbf{98.85}
                   & 95.76 / 97.65 / 98.07
                   & 89.60 / 92.60 / 94.70 \\
 & Append "Wait"             & 94.62 / 94.20 / 93.19
                   & 96.67 / \textbf{97.30} / \textbf{96.93}
                   & \textbf{100.00} / 99.30 / 98.39
                   & 90.15 / 94.47 / 96.33
                   & 85.20 / 89.20 / 93.30 \\
 & Original         & 96.31 / 95.65 / 92.93
                   & 93.89 / 96.22 / 95.60
                   & 99.52 / 99.30 / 97.70
                   & 89.39 / 94.32 / 96.25
                   & 86.40 / 91.40 / 93.50 \\
\midrule
\multirow{5}{*}{\textbf{deepseek-llama3-8B}}
 & ThinkEdit (8\%)  & 96.31 / \textbf{96.49} / \textbf{95.97}
                   & \textbf{97.78} / 97.57 / \textbf{98.40}
                   & 99.05 / \textbf{99.30} / 98.85
                   & \textbf{97.12} / \textbf{97.58} / 97.39
                   & 95.20 / \textbf{94.20} / 94.80 \\
 & ThinkEdit (4\%)  & 96.31 / 95.50 / 94.68
                   & \textbf{97.78} / 97.57 / 97.60
                   & 99.05 / 99.07 / 97.82
                   & 95.76 / 97.42 / \textbf{97.46}
                   & \textbf{95.60} / 93.80 / \textbf{95.40} \\
 & ThinkEdit (2\%)  & \textbf{97.08} / 95.27 / 93.95
                   & \textbf{97.78} / \textbf{98.65} / 97.87
                   & \textbf{100.00} / \textbf{99.30} / 98.62
                   & 95.61 / 96.89 / 97.12
                   & 92.80 / 93.60 / 94.40 \\
 & Append "Wait"             & 88.15 / 89.01 / 88.29
                   & \textbf{97.78} / 97.57 / 97.87
                   & 98.57 / 97.21 / 95.75
                   & 79.55 / 89.02 / 93.45
                   & 86.40 / 86.00 / 90.70 \\
 & Original         & 88.92 / 87.18 / 85.82
                   & 97.22 / 96.49 / 96.80
                   & 97.14 / 94.88 / 94.83
                   & 78.64 / 88.79 / 93.41
                   & 82.00 / 81.40 / 88.30 \\
\midrule
\multirow{5}{*}{\textbf{deepseek-qwen-1.5B}}
 & ThinkEdit (8\%)  & \textbf{95.38} / \textbf{94.20} / \textbf{92.97}
                   & \textbf{93.89} / \textbf{92.70} / \textbf{91.87}
                   & 94.76 / 96.05 / 96.90
                   & \textbf{96.21} / \textbf{97.20} / 96.78
                   & \textbf{94.00} / 93.60 / 94.40 \\
 & ThinkEdit (4\%)  & 92.62 / 92.90 / 92.32
                   & 87.78 / 88.11 / 88.67
                   & 95.71 / 95.58 / 96.44
                   & 95.15 / 96.59 / \textbf{97.27}
                   & 90.80 / 92.00 / 94.20 \\
 & ThinkEdit (2\%)  & 92.46 / 92.37 / 92.05
                   & 77.22 / 80.54 / 79.73
                   & 96.19 / 95.81 / \textbf{97.36}
                   & 93.79 / 95.83 / 95.80
                   & 92.80 / \textbf{94.40} / \textbf{94.90} \\
 & Append "Wait"             & 88.92 / 87.10 / 86.77
                   & 82.22 / 79.46 / 76.13
                   & 96.67 / \textbf{96.74} / 96.44
                   & 92.27 / 94.85 / 95.72
                   & 86.00 / 90.60 / 92.30 \\
 & Original         & 88.46 / 87.48 / 85.02
                   & 62.78 / 62.16 / 60.53
                   & \textbf{97.62} / 95.12 / 93.91
                   & 91.52 / 95.00 / 95.72
                   & 82.40 / 89.80 / 93.40 \\
\bottomrule
\end{tabular*}
\vspace{-10pt}
\caption{Accuracy (\%) on the top 5\% / 10\% / 20\% shortest responses for ThinkEdit with different editing rates.}
\label{tab:short_acc_detailed}
\end{table}
\setlength{\tabcolsep}{4pt}
\begin{table}[H]
\centering
\scriptsize
\begin{tabular*}{\linewidth}{@{\extracolsep{\fill}} l l c c c c c}
\toprule
\textbf{Model} & & \textbf{GSM8K} & \textbf{MMLU Elem.\ Math} & \textbf{MATH-Lvl1} & \textbf{MATH-Lvl5} & \textbf{MATH-500}\\
\midrule
\multirow{5}{*}{\textbf{deepseek-qwen-14B}}
 & ThinkEdit (8\%) & 113.2 / 129.4 / 153.6
                  & 86.9 / 99.0 / 117.2
                  & \textbf{180.7} / \textbf{238.5} / \textbf{372.3}
                  & \textbf{768.1} / 925.6 / 1136.0
                  & \textbf{414.7} / \textbf{573.9} / 759.0 \\
 & ThinkEdit (4\%) & 101.7 / 113.6 / 131.0
                  & 82.7 / 91.8 / 105.6
                  & 146.7 / 188.6 / 346.0
                  & 745.5 / \textbf{926.6} / \textbf{1163.7}
                  & 361.3 / 559.3 / \textbf{764.6} \\
 & ThinkEdit (2\%) & 95.4 / 106.3 / 120.2
                  & 79.1 / 87.1 / 98.7
                  & 125.1 / 150.2 / 243.4
                  & 698.5 / 906.6 / 1157.2
                  & 270.2 / 492.6 / 733.3 \\
 & Wait            & \textbf{127.2} / \textbf{145.0} / \textbf{166.0}
                  & \textbf{104.1} / \textbf{114.4} / \textbf{127.6}
                  & 159.3 / 191.8 / 281.9
                  & 672.1 / 875.5 / 1132.1
                  & 293.6 / 495.7 / 720.6 \\
 & Original        & 76.6 / 86.5 / 99.1
                  & 65.8 / 72.2 / 80.6
                  & 93.7 / 114.3 / 188.6
                  & 628.8 / 858.4 / 1125.9
                  & 198.7 / 434.3 / 697.0 \\
\midrule
\multirow{5}{*}{\textbf{deepseek-llama3-8B}}
 & ThinkEdit (8\%) & \textbf{160.4} / \textbf{185.7} / \textbf{225.2}
                  & 426.0 / 484.4 / 559.4
                  & \textbf{209.5} / \textbf{267.2} / \textbf{380.8}
                  & \textbf{825.3} / \textbf{978.8} / \textbf{1190.7}
                  & \textbf{422.6} / \textbf{567.4} / 759.5 \\
 & ThinkEdit (4\%) & 110.3 / 131.8 / 164.6
                  & 398.5 / 462.4 / 541.8
                  & 176.3 / 221.3 / 336.0
                  & 806.1 / 963.3 / 1185.1
                  & 372.5 / 553.2 / \textbf{772.9} \\
 & ThinkEdit (2\%) & 93.2 / 106.9 / 127.4
                  & 396.5 / 464.2 / 543.2
                  & 137.4 / 173.3 / 277.1
                  & 791.2 / 954.8 / 1185.1
                  & 305.2 / 506.3 / 737.6 \\
 & Wait            & 132.2 / 148.2 / 169.1
                  & \textbf{444.5} / \textbf{501.7} / \textbf{565.9}
                  & 148.4 / 179.2 / 244.0
                  & 680.8 / 887.3 / 1147.1
                  & 277.9 / 452.1 / 693.5 \\
 & Original        & 73.0 / 83.1 / 96.6
                  & 371.0 / 438.1 / 518.2
                  & 80.3 / 97.2 / 130.3
                  & 617.9 / 854.9 / 1126.5
                  & 159.5 / 357.5 / 644.5 \\
\midrule
\multirow{5}{*}{\textbf{deepseek-qwen-1.5B}}
 & ThinkEdit (8\%) & 115.9 / \textbf{138.2} / \textbf{180.1}
                  & 87.4 / 103.7 / \textbf{130.1}
                  & \textbf{247.3} / \textbf{396.1} / \textbf{577.3}
                  & \textbf{859.4} / 1003.7 / 1216.6
                  & \textbf{595.9} / \textbf{719.8} / \textbf{871.6} \\
 & ThinkEdit (4\%) & 103.3 / 118.9 / 144.8
                  & 80.6 / 92.5 / 112.9
                  & 172.7 / 336.9 / 543.6
                  & 853.0 / 1003.5 / 1221.9
                  & 530.8 / 676.0 / 837.4 \\
 & ThinkEdit (2\%) & 97.2 / 109.4 / 126.3
                  & 75.9 / 85.0 / 99.5
                  & 127.9 / 174.1 / 416.4
                  & 818.0 / 984.5 / 1214.3
                  & 435.0 / 612.9 / 800.6 \\
 & Wait            & \textbf{120.6} / 137.0 / 158.0
                  & \textbf{101.6} / \textbf{112.9} / 128.0
                  & 147.9 / 180.1 / 310.2
                  & 822.7 / \textbf{1020.9} / 1306.0
                  & 341.8 / 556.6 / 791.8 \\
 & Original        & 78.8 / 89.4 / 103.0
                  & 61.6 / 68.5 / 77.6
                  & 88.8 / 110.3 / 219.7
                  & 804.6 / 1017.9 / \textbf{1314.0}
                  & 249.7 / 506.5 / 760.7 \\
\bottomrule
\end{tabular*}
\vspace{-10pt}
\caption{Average reasoning length (in tokens) of the top 5\% / 10\% / 20\% shortest responses for ThinkEdit with different editing rates.}
\label{tab:short_len_detailed}
\end{table}

\clearpage

\subsection{ThinkEdit Results for 32B Reasoning Model}
\label{sec:32b_results}
We report results for the larger \texttt{deepseek-distill-qwen-32B} model. Although ThinkEdit does not yield overall accuracy gains on the MATH-series datasets (Table~\ref{tab:32b_overall_acc}), it consistently achieves higher accuracy on short reasoning responses similar to the smaller models (Table~\ref{tab:32b_short_reasoning_acc}).

\begin{table}[H]
\centering
\scriptsize
\begin{tabular*}{\linewidth}{@{\extracolsep{\fill}} l l c c c c c}
\toprule
\textbf{Model} & & \textbf{GSM8K} & \textbf{MMLU Elem.\ Math} & \textbf{MATH-Level1} & \textbf{MATH-Level5} & \textbf{MATH-500}\\
\midrule
\multirow{5}{*}{\textbf{deepseek-qwen-32B}}
 & ThinkEdit (8\%) & $\mathbf{95.34 \pm 0.23}$ & $\mathbf{98.10 \pm 0.17}$ & $95.31 \pm 0.38$ & $91.16 \pm 0.45$ & $91.44 \pm 0.57$\\
 & ThinkEdit (4\%) & $95.25 \pm 0.25$ & $98.02 \pm 0.31$ & $96.02 \pm 0.42$ & $91.31 \pm 0.50$ & $91.60 \pm 0.65$\\
 & ThinkEdit (2\%) & $94.90 \pm 0.34$ & $97.49 \pm 0.49$ & $96.27 \pm 0.54$ & $91.31 \pm 0.29$ & $\mathbf{91.62 \pm 0.74}$\\
 & Append "Wait"        & $92.72 \pm 0.54$ & $96.16 \pm 0.45$ & $96.27 \pm 0.39$ & $\mathbf{91.32 \pm 0.46}$ & $91.46 \pm 0.51$\\
 & Original        & $92.97 \pm 0.39$ & $95.93 \pm 0.83$ & $\mathbf{96.41 \pm 0.45}$ & $91.27 \pm 0.53$ & $\mathbf{91.62 \pm 0.58}$\\
\bottomrule
\end{tabular*}
\vspace{-10pt}
\caption{Overall accuracy (\%) of \texttt{deepseek-distill-qwen-32B} with different ThinkEdit edit-rates.}
\label{tab:32b_overall_acc}
\end{table}
\setlength{\tabcolsep}{4pt}
\begin{table}[H]
\centering
\scriptsize
\begin{tabular*}{\linewidth}{@{\extracolsep{\fill}} l l c c c c c}
\toprule
\textbf{Model} & & \textbf{GSM8K} & \textbf{MMLU Elem.\ Math} & \textbf{MATH-Level1} & \textbf{MATH-Level5} & \textbf{MATH-500}\\
\midrule
\multirow{5}{*}{\textbf{deepseek-qwen-32B}}
 & ThinkEdit (8\%) & $\mathbf{665.6 \pm 762.8}$ & $\mathbf{312.3 \pm 332.0}$ & $\mathbf{1548.6 \pm 1473.4}$ & $3676.7 \pm 3388.7$ & $2665.6 \pm 2885.1$\\
 & ThinkEdit (4\%) & $445.8 \pm 684.7$ & $287.7 \pm 600.0$ & $1484.7 \pm 1587.7$ & $3821.1 \pm 3518.3$ & $2736.4 \pm 2948.8$\\
 & ThinkEdit (2\%) & $405.3 \pm 620.5$ & $238.8 \pm 315.9$ & $1485.3 \pm 1622.1$ & $3947.0 \pm 3564.7$ & $2816.1 \pm 3029.2$\\
 & Append "Wait"        & $405.5 \pm 519.0$ & $280.6 \pm 401.5$ & $1484.8 \pm 1619.1$ & $\mathbf{4103.9 \pm 3606.0}$ & $\mathbf{2878.8 \pm 3029.3}$\\
 & Original        & $306.2 \pm 515.4$ & $168.9 \pm 105.3$ & $1457.6 \pm 1621.0$ & $4100.7 \pm 3608.7$ & $2872.0 \pm 3034.8$\\
\bottomrule
\end{tabular*}
\vspace{-10pt}
\caption{Overall reasoning length (in tokens) for \texttt{deepseek-distill-qwen-32B}.}
\label{tab:32b_overall_len}
\end{table}
\begin{table}[H]
\centering
\setlength{\tabcolsep}{4pt} 
\scriptsize
\begin{tabular*}{\linewidth}{@{\extracolsep{\fill}} l l c c c c c}
\toprule
\textbf{Model} & & \textbf{GSM8K} & \textbf{MMLU Elem. Math} & \textbf{MATH-Level1} & \textbf{MATH-Level5} & \textbf{MATH-500} \\
\midrule
\multirow{5}{*}{\textbf{deepseek-qwen-32B}}
 & ThinkEdit (8\%) 
    & \textbf{99.08} / \textbf{98.47} / \textbf{97.95} 
    & \textbf{98.33} / \textbf{97.57} / 97.07 
    & 99.52 / 98.60 / 97.36 
    & \textbf{99.55} / \textbf{99.39} / \textbf{98.64} 
    & \textbf{94.40} / \textbf{95.40} / \textbf{96.10} \\
 & ThinkEdit (4\%) 
    & 98.92 / 97.71 / 97.83 
    & 97.78 / \textbf{97.57} / \textbf{97.20} 
    & \textbf{100.00} / \textbf{100.00} / 98.74 
    & 98.03 / 98.64 / 97.99 
    & 92.00 / 94.40 / 95.80 \\
 & ThinkEdit (2\%) 
    & 98.92 / 98.24 / 97.68 
    & 96.67 / 97.03 / 96.80 
    & 99.05 / 98.84 / 98.51 
    & 97.58 / 98.26 / 98.22 
    & 90.00 / 92.60 / 94.70 \\
 & Append "Wait"            
    & 97.08 / 96.03 / 95.21 
    & 95.00 / 96.76 / 96.27 
    & 99.52 / 99.30 / 98.05 
    & 94.09 / 96.89 / 97.61 
    & 84.80 / 90.40 / 93.20 \\
 & Original        
    & 98.31 / 97.18 / 96.20 
    & 97.78 / 97.03 / 95.87 
    & \textbf{100.00} / \textbf{100.00} / \textbf{98.97} 
    & 93.03 / 96.36 / 97.35 
    & 86.40 / 92.00 / 94.00 \\
\bottomrule
\end{tabular*}
\vspace{-10pt}
\caption{Accuracy (\%) on the top 5\% / 10\% / 20\% shortest responses for \texttt{deepseek-distill-qwen-32B}.}
\label{tab:32b_short_reasoning_acc}
\end{table}

\setlength{\tabcolsep}{4pt}
\begin{table}[H]
\centering
\scriptsize
\begin{tabular*}{\linewidth}{@{\extracolsep{\fill}} l l c c c c c}
\toprule
\textbf{Model} & & \textbf{GSM8K} & \textbf{MMLU Elem.\ Math} & \textbf{MATH-Lvl1} & \textbf{MATH-Lvl5} & \textbf{MATH-500} \\
\midrule
\multirow{5}{*}{\textbf{deepseek-qwen-32B}}
 & ThinkEdit (8\%) 
    & 105.2 / 121.8 / 148.6 
    & 89.2 / 100.5 / 117.7 
    & \textbf{367.8} / \textbf{492.8} / \textbf{625.4} 
    & \textbf{793.5} / \textbf{919.5} / 1094.6 
    & \textbf{567.1} / \textbf{677.0} / \textbf{811.1} \\
 & ThinkEdit (4\%) 
    & 95.2 / 105.8 / 120.1  
    & 85.9 / 96.1 / 110.6  
    & 146.9 / 202.2 / 360.9 
    & 751.1 / 905.4 / 1101.0 
    & 446.7 / 600.0 / 768.9 \\
 & ThinkEdit (2\%) 
    & 93.2 / 103.6 / 116.6  
    & 79.1 / 88.6 / 101.5  
    & 124.3 / 155.3 / 307.6 
    & 746.4 / 910.8 / 1123.7 
    & 371.3 / 563.0 / 759.8 \\
 & Append "Wait"            
    & \textbf{125.7} / \textbf{143.0} / \textbf{163.7} 
    & \textbf{109.6} / \textbf{121.1} / \textbf{135.9} 
    & 151.4 / 182.0 / 247.2 
    & 725.7 / 914.4 / \textbf{1153.4} 
    & 328.4 / 521.3 / 739.4 \\
 & Original        
    & 76.7 / 86.7 / 99.6    
    & 69.3 / 76.1 / 84.3   
    & 89.9 / 109.4 / 149.6 
    & 672.7 / 886.7 / 1139.2 
    & 216.4 / 454.9 / 705.9 \\
\bottomrule
\end{tabular*}
\vspace{-10pt}
\caption{Average reasoning length (tokens) of the top 5\% / 10\% / 20\% shortest responses for \texttt{deepseek-distill-qwen-32B}.}
\label{tab:32b_short_reasoning_len}
\end{table}

\clearpage

\subsection{ThinkEdit Results on Non-Math Domains}
\label{app:non_math_mmlu}
We include supplementary experiments on three non-math subjects from MMLU—\emph{High School Computer Science}, \emph{Formal Logic}, and \emph{Professional Accounting}. These tasks do not directly require mathematical calculation, but they still demand structured reasoning and logical consistency. As shown in Tables~\ref{tab:app_mmlu_overall} and \ref{tab:app_mmlu_short}, \textbf{\textit{ThinkEdit}} not only improves overall accuracy but also mitigates failure cases that arise from overly short reasoning traces. This indicates that \textbf{\textit{ThinkEdit}} can boost performance in reasoning-heavy domains beyond mathematics.

\begin{table}[H]
\centering
\scriptsize
\begin{tabular*}{\linewidth}{@{\extracolsep{\fill}} l l c c c}
\toprule
\textbf{Model} & \textbf{Variant} & \textbf{MMLU HS CS} & \textbf{MMLU Formal Logic} & \textbf{MMLU Prof. Accounting} \\
\midrule
\multirow{2}{*}{\textbf{deepseek-qwen-32B}}
& Original        & 96.50 $\pm$ 1.18 & 93.49 $\pm$ 1.29 & \textbf{84.22 $\pm$ 1.05} \\
& ThinkEdit (4\%) & \textbf{97.20 $\pm$ 0.92} & \textbf{94.05 $\pm$ 1.80} & 83.40 $\pm$ 1.89 \\
\midrule
\multirow{2}{*}{\textbf{deepseek-qwen-14B}}
& Original        & 93.50 $\pm$ 1.58 & 91.27 $\pm$ 2.08 & 75.85 $\pm$ 1.89 \\
& ThinkEdit (4\%) & \textbf{94.80 $\pm$ 1.14} & \textbf{91.51 $\pm$ 1.98} & \textbf{77.09 $\pm$ 1.63} \\
\midrule
\multirow{2}{*}{\textbf{deepseek-llama3-8B}}
& Original        & 84.00 $\pm$ 3.83 & 63.41 $\pm$ 1.47 & 57.06 $\pm$ 1.48 \\
& ThinkEdit (4\%) & \textbf{88.90 $\pm$ 1.91} & \textbf{65.40 $\pm$ 2.85} & \textbf{57.52 $\pm$ 1.40} \\
\midrule
\multirow{2}{*}{\textbf{deepseek-qwen-1.5B}}
& Original        & 63.90 $\pm$ 4.38 & 51.27 $\pm$ 3.00 & 35.71 $\pm$ 2.85 \\
& ThinkEdit (4\%) & \textbf{68.30 $\pm$ 3.02} & \textbf{52.30 $\pm$ 3.07} & \textbf{37.09 $\pm$ 1.78} \\
\bottomrule
\end{tabular*}
\vspace{-10pt}
\caption{Overall accuracy on MMLU non-math subjects.}
\label{tab:app_mmlu_overall}
\end{table}
\begin{table}[H]
\centering
\scriptsize
\begin{tabular*}{\linewidth}{@{\extracolsep{\fill}} l l c c c}
\toprule
\textbf{Model} & \textbf{Variant} & \textbf{MMLU HS CS} & \textbf{MMLU Formal Logic} & \textbf{MMLU Prof. Accounting} \\
\midrule
\multirow{2}{*}{\textbf{deepseek-qwen-32B}}
& Original        & 98.00 / 98.00 / 97.50 & 85.00 / \textbf{88.33} / 91.60 & 89.29 / 87.86 / 87.86 \\
& ThinkEdit (4\%) & \textbf{100.00} / \textbf{99.00} / \textbf{99.50} & \textbf{88.33} / \textbf{88.33} / \textbf{92.00} & \textbf{92.86} / \textbf{91.07} / \textbf{91.96} \\
\midrule
\multirow{2}{*}{\textbf{deepseek-qwen-14B}}
& Original        & 96.00 / 94.00 / 96.00 & 78.33 / 85.83 / 90.40 & 82.14 / 82.50 / 84.82 \\
& ThinkEdit (4\%) & \textbf{100.00} / \textbf{99.00} / \textbf{99.50} & \textbf{85.00} / \textbf{90.00} / \textbf{92.00} & \textbf{90.00} / \textbf{91.43} / \textbf{90.36} \\
\midrule
\multirow{2}{*}{\textbf{deepseek-llama3-8B}}
& Original        & 72.00 / 76.00 / 81.00 & 75.00 / 75.83 / 74.00 & 70.71 / 70.71 / 67.32 \\
& ThinkEdit (4\%) & \textbf{96.00} / \textbf{95.00} / \textbf{96.00} & \textbf{80.00} / \textbf{77.50} / \textbf{81.20} & \textbf{75.00} / \textbf{74.64} / \textbf{70.54} \\
\midrule
\multirow{2}{*}{\textbf{deepseek-qwen-1.5B}}
& Original        & 66.00 / 66.00 / 71.00 & 48.33 / 51.67 / 61.60 & 32.86 / 32.14 / 33.21 \\
& ThinkEdit (4\%) & \textbf{92.00} / \textbf{89.00} / \textbf{84.00} & \textbf{60.00} / \textbf{61.67} / \textbf{67.20} & \textbf{43.57} / \textbf{43.21} / \textbf{40.36} \\
\bottomrule
\end{tabular*}
\vspace{-10pt}
\caption{Accuracy on the top 5\% / 10\% / 20\% shortest responses on MMLU non-math subjects.}
\label{tab:app_mmlu_short}
\end{table}

\clearpage

\subsection{Behavioral Analysis: ThinkEdit Encourages Deeper Reasoning}
\label{app:behavioral_analysis}
To better understand how \textbf{\textit{ThinkEdit}} influences reasoning quality beyond final correctness, we analyze model behavior over \emph{all} examples (correct and incorrect). We quantify (i) the average number of LaTeX equations written in the \texttt{<think>} trace, (ii) the percentage of examples whose final answer is explicitly boxed in \texttt{<think>}, and (iii) the average reasoning length (tokens) within \texttt{<think>}. 

Across the four base models, \textbf{\textit{ThinkEdit}} consistently produces more equations (+55.2\% on average), boxes the final answer more often (+60.5\%), and writes longer reasoning traces (+48.0\%) compared to the original models. The effect is especially pronounced for smaller models (e.g., \texttt{deepseek-qwen-1.5B}: equations +83.9\%, boxing +84.4\%). These results indicate that the edits reliably encourage more explicit and structured chains of thought across the board.

\begin{table}[H]
\centering
\scriptsize
\begin{tabular*}{\linewidth}{@{\extracolsep{\fill}} l l c c c}
\toprule
\textbf{Model} & \textbf{Variant} & \textbf{Equations in CoT} & \textbf{Wrap Answer in CoT (\%)} & \textbf{Reasoning Length (tokens)} \\
\midrule
\multirow{2}{*}{\textbf{deepseek-qwen-32B}}
& Original        & 1.92 & 11.75 & 308.32 \\
& ThinkEdit (4\%) & \textbf{2.85} & \textbf{16.00} & \textbf{409.03} \\
\midrule
\multirow{2}{*}{\textbf{deepseek-qwen-14B}}
& Original        & 2.11 & 15.39 & 341.09 \\
& ThinkEdit (4\%) & \textbf{2.74} & \textbf{23.73} & \textbf{518.06} \\
\midrule
\multirow{2}{*}{\textbf{deepseek-llama3-8B}}
& Original        & 3.21 & 30.25 & 606.09 \\
& ThinkEdit (4\%) & \textbf{5.09} & \textbf{50.57} & \textbf{957.32} \\
\midrule
\multirow{2}{*}{\textbf{deepseek-qwen-1.5B}}
& Original        & 3.83 & 28.05 & 753.45 \\
& ThinkEdit (4\%) & \textbf{7.05} & \textbf{51.71} & \textbf{1125.64} \\
\bottomrule
\end{tabular*}
\vspace{-8pt}
\caption{Behavioral comparison over all datasets. \textbf{\textit{ThinkEdit}} generates more equations, is more likely to box the final answer, and produces longer reasoning traces than the original models, indicating more explicit and structured CoT across cases.}
\label{tab:app_error_analysis}
\end{table}

\clearpage

\subsection{Examples of Steering the Reasoning Length}
\label{sec:steering_examples}

\begin{figure}[H]
    \centering
    \includegraphics[width=1.0\textwidth]{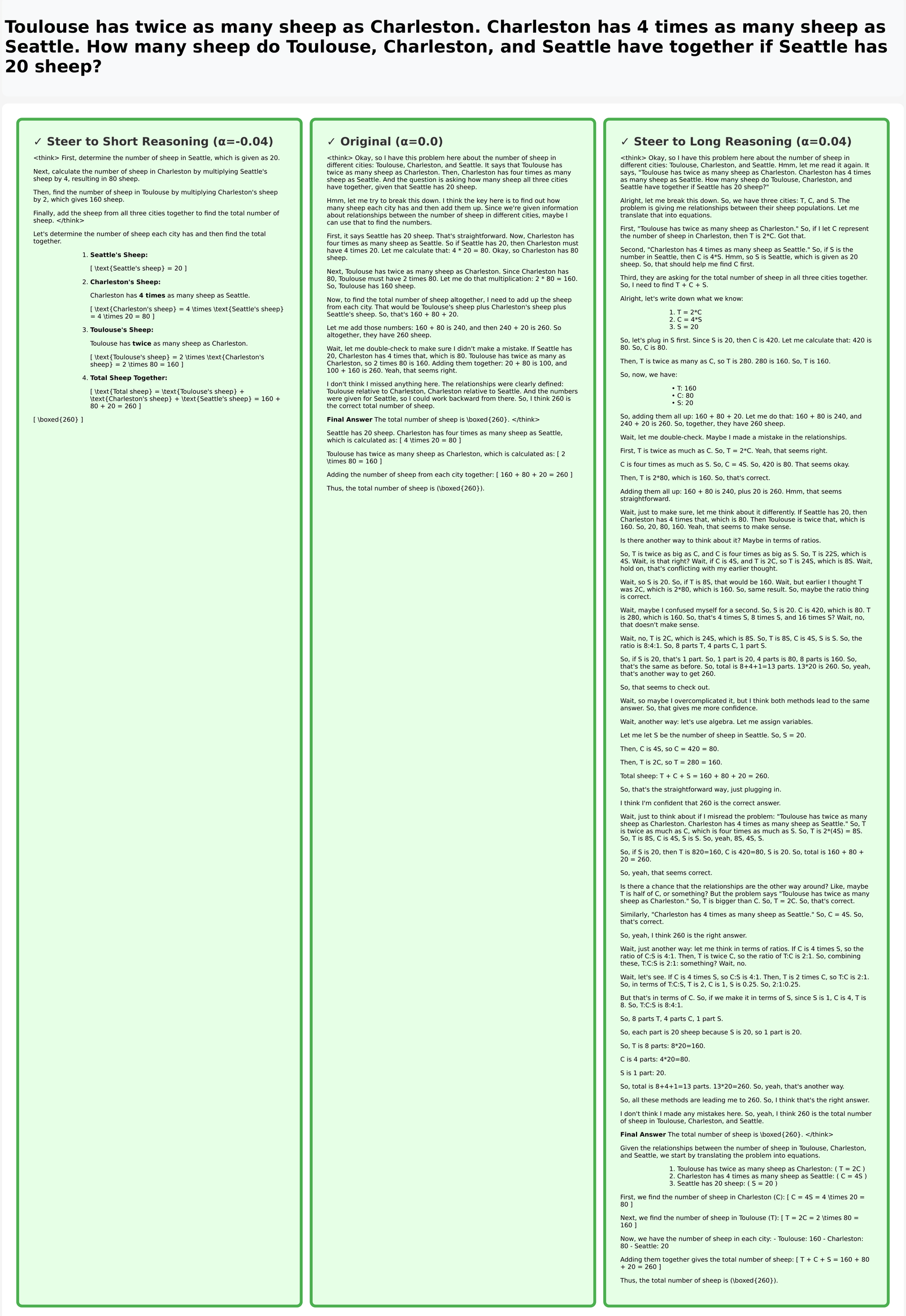}
    \vspace{-20pt}
    \caption{GSM8k Example 1: Steering the reasoning length of deepseek-distill-llama3-8b. Left: shorter reasoning ($\alpha = -0.04$), Middle: original response, Right: longer reasoning ($\alpha = 0.04$).}
    \label{fig:steering_example_1}
\end{figure}

\begin{figure}[H]
    \centering
    \includegraphics[width=1.0\textwidth]{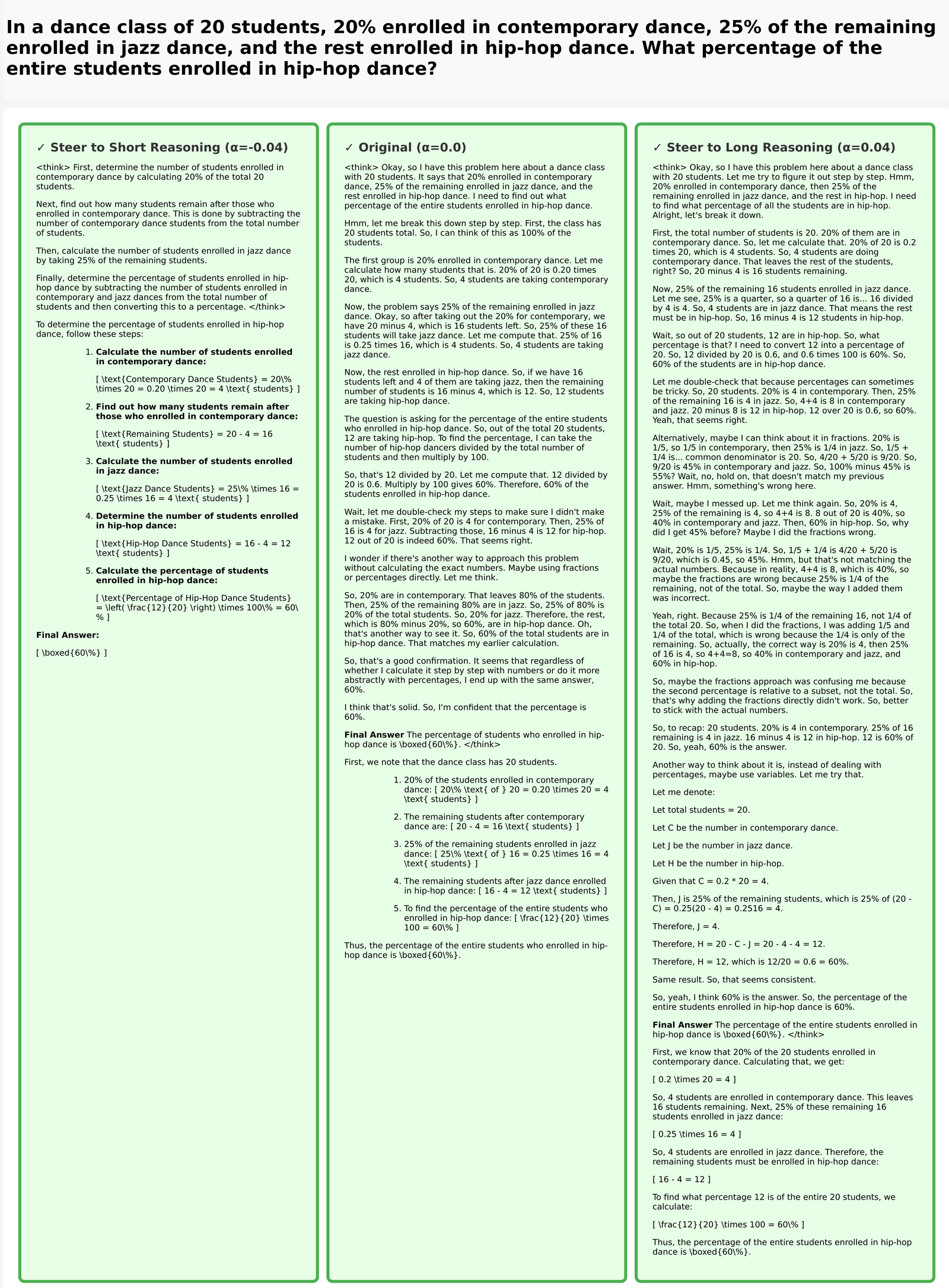}
    \vspace{-20pt}
    \caption{GSM8k Example 2: Steering the reasoning length of deepseek-distill-llama3-8b. Left: shorter reasoning ($\alpha = -0.04$), Middle: original response, Right: longer reasoning ($\alpha = 0.04$).}
    \label{fig:steering_example_2}
\end{figure}

To demonstrate the effect of steering the reasoning length, we present two examples from gsm8k in Figures~\ref{fig:steering_example_1} and~\ref{fig:steering_example_2}. Each figure contains three different reasoning outputs:

\begin{itemize}[leftmargin=1.5em]
    \item \textbf{Left}: The model's response when steered towards shorter reasoning with $\alpha = -0.04$.
    \item \textbf{Middle}: The original unaltered response.
    \item \textbf{Right}: The model's response when steered towards longer reasoning with $\alpha = 0.04$.
\end{itemize}

These examples highlight that steering along the reasoning direction effectively modulates the reasoning length without causing unintended truncation or artificial elongation. In contrast to naive methods such as forcibly stopping the reasoning process—resulting in incomplete reasoning—or appending redundant tokens like "Wait" to artificially extend responses, our approach enables the model to generate complete and coherent reasoning of varying lengths. By directly manipulating the internal representation, we achieve a more controlled and flexible mechanism for managing reasoning length, which could be particularly useful for constrained computational budgets.

\clearpage

\clearpage

\subsection{Examples of Fixing the Overly Short Reasoning with \textit{ThinkEdit}}
\label{sec:examples_editing}

To illustrate the effectiveness of our \textbf{\textit{ThinkEdit}} approach in addressing the overly short reasoning issue, we show two representative examples from the GSM8K and MATH-level5 datasets.

\begin{figure}[H]
    \centering
    \includegraphics[width=1.0\linewidth]{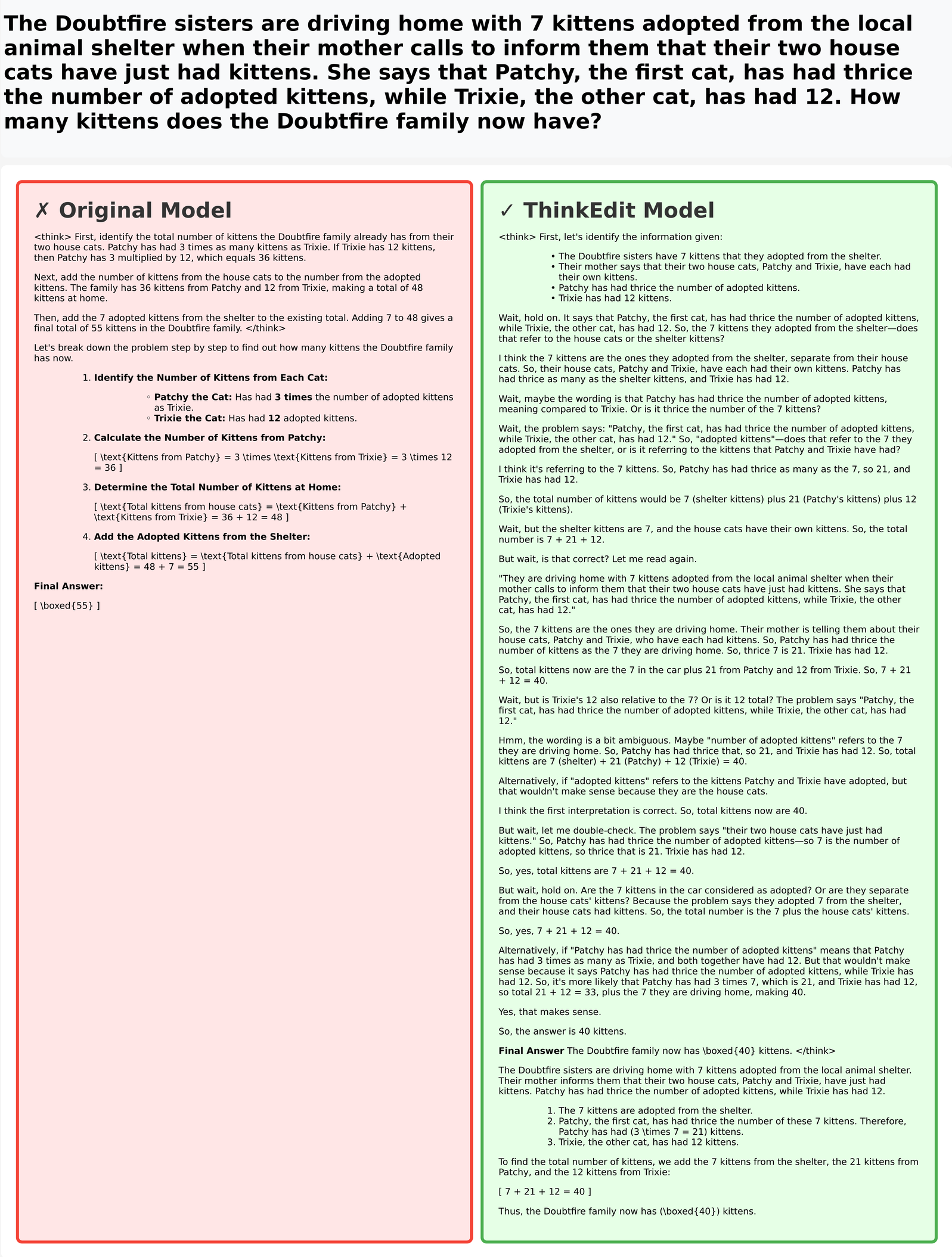}
    \caption{An example of \textbf{\textit{ThinkEdit}} from the GSM8K dataset. The original model provides a short, flawed explanation. After \textbf{\textit{ThinkEdit}}, the model’s reasoning process is more thorough.}
    \label{fig:editing_example_gsm8k}
\end{figure}

\paragraph{Example from \textsc{gsm8k}.}
The example is shown in Figure~\ref{fig:editing_example_gsm8k}. In this problem, the \emph{Original} model misinterprets the question and quickly provides an incorrect conclusion. In contrast, after applying \textbf{\textit{ThinkEdit}}, the model first organizes the given information and displays a detailed reasoning process that leads to the correct answer.

\begin{figure}[H]
    \centering
    \includegraphics[width=1.0\linewidth]{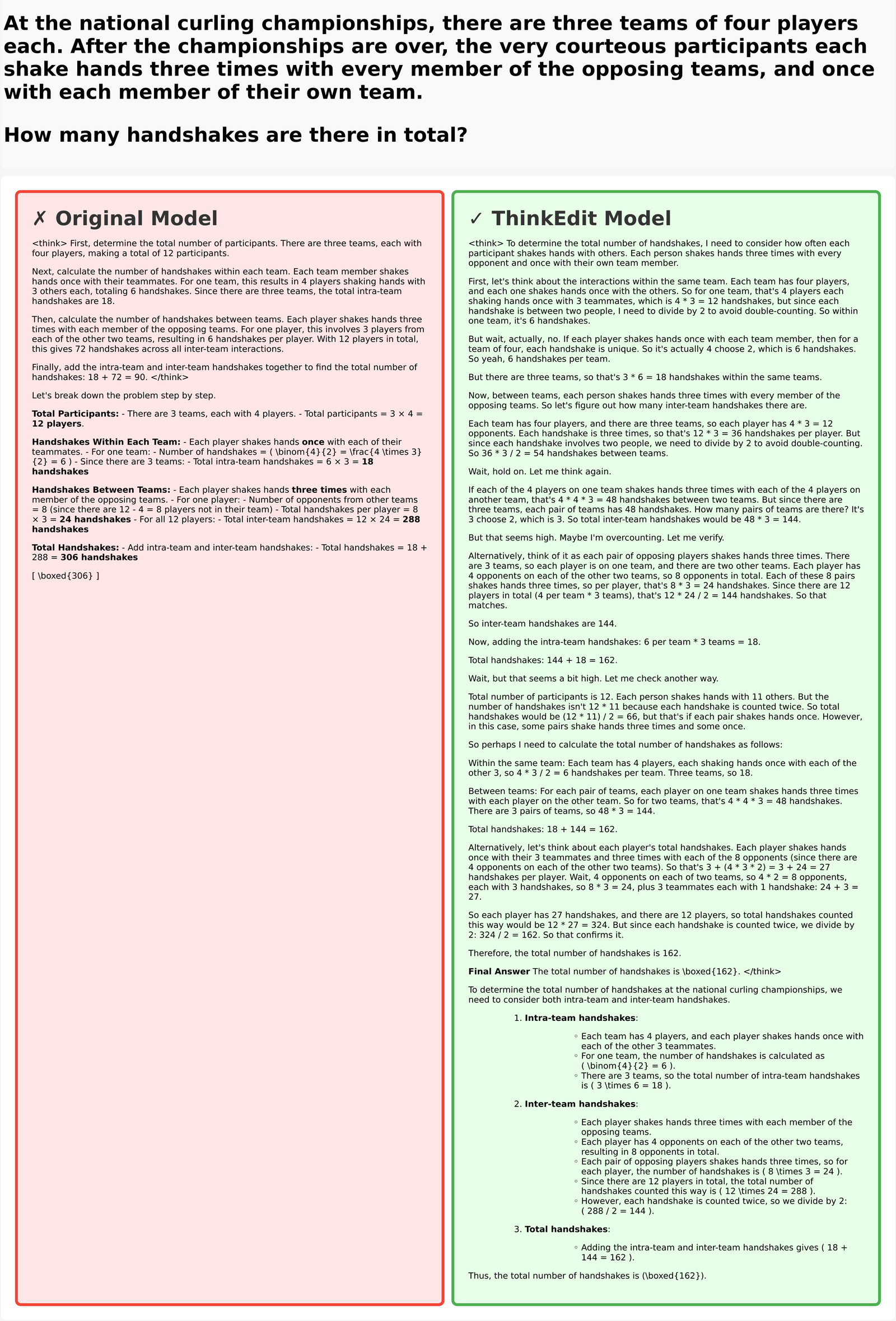}
    \caption{An example of \textbf{\textit{ThinkEdit}} from MATH-level5. While both models initially make a wrong assumption, the model after applying \textbf{\textit{ThinkEdit}} corrects itself and arrives at the correct final reasoning.}
    \label{fig:editing_example_MATH}
\end{figure}

\paragraph{Example from \textsc{MATH-level5}.} The example is shown in Figure~\ref{fig:editing_example_MATH}. Both the \emph{Original} model and the model with \textbf{\textit{ThinkEdit}} begin with an incorrect line of reasoning. However, the original model does not revise its approach and proceeds to an incorrect final conclusion. In contrast, the model with \textbf{\textit{ThinkEdit}} revisits its reasoning steps, recognizes the mistake, and corrects the process in time to reach the correct solution.

These examples demonstrate \textbf{\textit{ThinkEdit}} can facilitate deeper, more accurate reasoning by guiding the model to properly utilize problem details, reconsider faulty steps, and ultimately yield correct solutions.

\clearpage

\end{document}